\documentclass{article}

\PassOptionsToPackage{numbers,sort&compress}{natbib}

\usepackage[preprint]{neurips_2025}
\usepackage{graphicx}
\usepackage{tikz}
\usepackage[export]{adjustbox}
\usepackage{caption}
\usepackage{subcaption}
\usetikzlibrary{calc}

\usepackage[utf8]{inputenc} 
\usepackage[T1]{fontenc}    
\usepackage{hyperref}       
\usepackage{url}            
\usepackage{booktabs}       
\usepackage{amsfonts}       
\usepackage{amsmath}        
\usepackage{nicefrac}       
\usepackage{microtype}      
\usepackage{xcolor}         

\title{SIEDD: Shared‑Implicit Encoder with Discrete Decoders}

\author{
  Vikram Rangarajan \\
  University of Maryland\\
  \texttt{vikramr@terpmail.umd.edu}\\
  \And
  Shishira R Maiya \\
  University of Maryland\\
  \texttt{shishira@umd.edu}
  \And
  Max Ehrlich \\
  University of Maryland\\
  \texttt{maxehr@umd.edu}\\
  \And
  Abhinav Shrivastava \\
  University of Maryland\\
  \texttt{abhinav@cs.umd.edu}
}

\begin{document}

\maketitle

\begin{abstract}
Implicit Neural Representations (INRs) offer exceptional fidelity for video compression by learning per-video optimized functions, but their adoption is crippled by impractically slow encoding times. Existing attempts to accelerate INR encoding often sacrifice reconstruction quality or crucial coordinate-level control essential for adaptive streaming and transcoding. We introduce SIEDD (Shared-Implicit Encoder with Discrete Decoders), a novel architecture that fundamentally accelerates INR encoding without these compromises. SIEDD first rapidly trains a shared, coordinate-based encoder on sparse anchor frames to efficiently capture global, low-frequency video features. This encoder is then frozen, enabling massively parallel training of lightweight, discrete decoders for individual frame groups, further expedited by aggressive coordinate-space sampling. This synergistic design delivers a remarkable 20-30X encoding speed-up over state-of-the-art INR codecs on HD and 4K benchmarks, while maintaining competitive reconstruction quality and compression ratios. Critically, SIEDD retains full coordinate-based control, enabling continuous resolution decoding and eliminating costly transcoding. Our approach significantly advances the practicality of high-fidelity neural video compression, demonstrating a scalable and efficient path towards real-world deployment. Our codebase is available at \url{https://github.com/VikramRangarajan/SIEDD}.
\end{abstract}

\section{Introduction}


Video data forms the majority of internet traffic and it is projected to grow exponentially over the next decade. Traditional video codecs \cite{sullivan2004overview,sullivan2012overview,bross2021overview} have hit a wall and the field is increasingly looking towards neural-based methods [cite] to deliver efficient rate-distortion trade-offs. Implicit Neural Representations (INRs) for videos offer an alternative functional representation of videos. Video-INRs have good compression and great decoding speeds, but suffer from slow encoding speeds, which makes them impractical.  

Unlike autoencoder-based video coding methods  \cite{li2021deep,li2024neural,jia2025towards}, Video-INRs are optimized per video, making them more truthful to the source, without any hallucinations \cite{zhang2021out} a necessary property. But this presents a huge problem - encoding a video is no longer a simple forward pass, but an extensive gradient optimization process which can take hours to encode a single clip. We stress on the fact that encoding time is crucial for widespread adoption of INR-based video codecs. For example, the price of encoding a single minute of 1080p video at 30fps costs around \$0.04 on AWS Mediaconvert,
while training a Video-INR like \cite{chen2023hnerv} on the same clip with an RTXA5000 would cost upwards of \$3,
a whopping 75x increase.  

Existing works \cite{strumpler2022implicit,lee2021meta,chen2024fast} in the field point towards having a good prior/initialization to be the key factor in imporving optimization times. However, these methods require huge memory \cite{lee2021meta} or do not scale beyond small video resolutions \cite{chen2024fast}, limiting their impact.
To overcome these limitations, we introduce SIEDD, a shared-encoder architecture designed for scalable and efficient encoding. We first rapidly train a shared encoder on a small set of keyframes—without requiring full convergence—to capture low-frequency, video-specific features. Inspired by findings in \cite{vyas2025learningtransferablefeaturesimplicit,kim2023generalizable}, we leverage the insight that early INR layers encode generalizable representations that converge quickly and transfer well across frames.
In contrast to frame-wise video INRs, which require per-frame encoding and lack spatial flexibility, our method takes normalized 2D coordinates as input. This enables continuous-resolution decoding from a single encoding pass—eliminating the need for resolution-specific transcoding and significantly reducing overhead. Once the encoder is trained, we freeze it and train lightweight, frame-group-specific decoders independently. This design enables scaling to long videos and allows parallelized decoder training, as demonstrated in \cite{maiya2023nirvana}. Additionally, by exploiting spatial sparsity, we subsample the coordinate space during training—yielding large gains in encoding speed without compromising reconstruction fidelity. Finally, by using simple MLP layers throughout, our architecture remains compatible with recent advances in LLM quantization \cite{badri2023hqq,dettmers2022llmint8}, enabling further compression without any architectural changes. SIEDD achieves an impressive $20\times$ encoding speed-up on UVG-HD \cite{10.1145/3339825.3394937} and over $30\times$ on UVG-4K \cite{10.1145/3339825.3394937}, while preserving high reconstruction quality. This speedup is a step towards making INR based video codecs more practical for deployment.

To summarize, our contributions are as follows:
\begin{itemize}
    \item A novel architecture with shared encoder and discrete decoders 
    that greatly speeds improves video encoding times of Video-INRs. Our model can scale both spatially (to 4K) and temporally (for longer videos) without any modifications. 
    \item A two-stage training process that uses the fact that early INR layers require fewer iterations to converge, combined with sparse sampling. 
    \item Extensive experiments on UVG \cite{10.1145/3339825.3394937}, UVG-4K and DAVIS datasets along with architectural ablations. 
\end{itemize} 
\section{Related Works}

\subsection{Video Compression}
Legacy video codecs like H264 \cite{sullivan2004overview}, HEVC \cite{sullivan2012overview} and the more recent VVC \cite{bross2021overview} operate on similar first principles. They compress videos by exploiting redundancy and motion between frames. However, such hand-engineered techniques and heuristics have hit a limit in terms of performance gains \cite{teng2024benchmarking}. Neural video codecs \cite{li2021deep,li2024neural,jia2025towards} build on existing autoencoder based hyper-prior architectures \cite{balle2018variational} to improve compression.

\subsection{Implicit Neural Representations}
Implicit Neural Representations (INRs). have emerged as a compact and differentiable paradigm for modeling continuous signals such as images\cite{sitzmann2020implicit}, videos \cite{chen2021nerv}, audio \cite{li2023aenerf} and 3D scenes \cite{mildenhall2020nerf}. Rather than storing discrete data, INRs encode a signal as the weights of a neural network that maps input coordinates to output values, offering high fidelity and resolution-agnostic reconstructions. Early works like SIREN~\cite{sitzmann2020implicit} demonstrated the expressivity of periodic activation functions in fitting detailed signals from scratch. Extending to video, NeRV~\cite{chen2021nerv} introduced a frame-wise INR architecture mapping timestamps to RGB frames, enabling fast inference but at the cost of limited spatial control. Subsequent efforts\cite{chen2023hnerv,maiya2023nirvana,kwan2023hinerv,lee2023ffnerv} addressed these limitations: HNeRV\cite{chen2023hnerv} introduced content-adaptive embeddings to improve convergence and generalization, while NIRVANA~\cite{maiya2023nirvana} adopted an autoregressive, patch-wise approach to exploit spatio-temporal redundancy and support scalable encoding of high-resolution, long-duration videos. Works like Tree-Nerv \cite{zhao2025tree}, DS-Nerv \cite{yan2024dsnerv}incorporated ideas of efficient sampling and dynamic codes to further improve these systems. 

\section{Model Compression}
With Video-INRs, the task of Video compression is essentially transformed into a model compression problem. In this functional paradigm, the challenge then becomes to effectively quantize \cite{han2016deep,frankle2019lottery,jacob2018quantization} and store the weights of the resulting neural network with minimal loss. We employ HQQ Quantization \cite{badri2023hqq} - a post training quantization technique combined with lossless entropy coding \cite{huffman1952method} to provide efficient bitstream.

\section{Method}

\subsection{Overview}

\begin{figure}
    \centering
    \includegraphics[width=\linewidth]{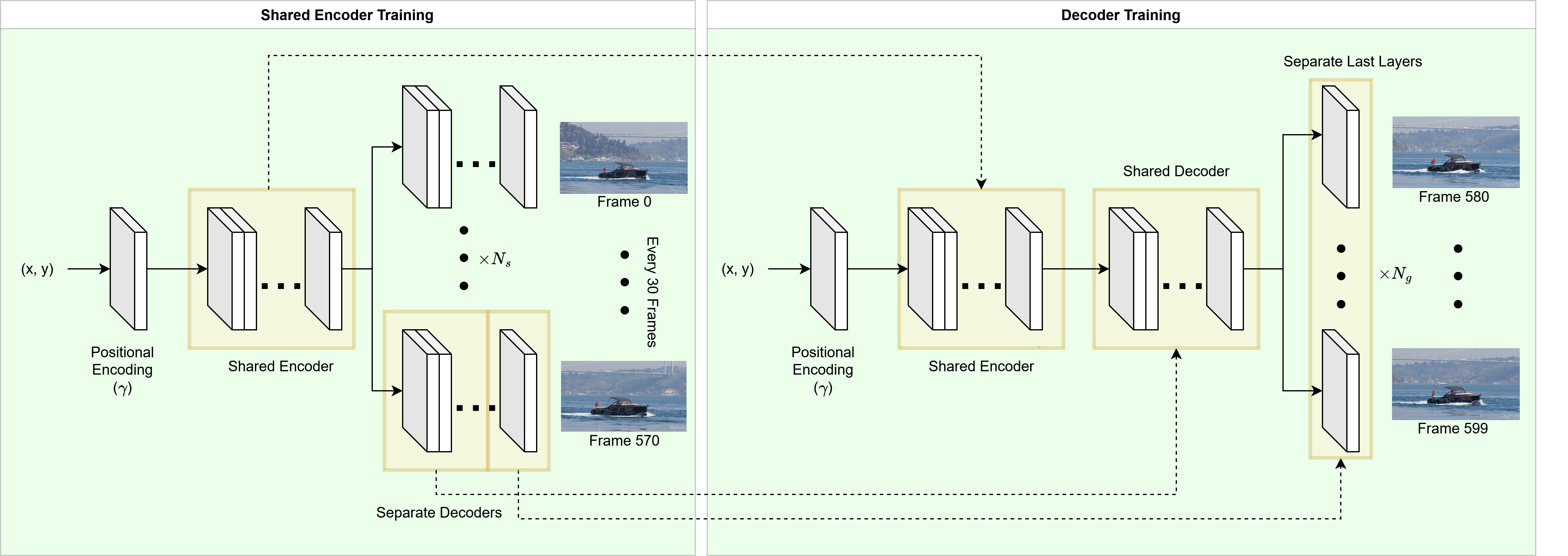}
    \caption{Overview of the SIEDD architecture. During the shared encoder training phase (left), a positional encoding of 2D coordinates is passed through a shared encoder and used to train a small number of frame-specific decoders on anchor frames sampled every \(N_g\) frames. In the decoder training phase (right), the encoder is frozen, and separate lightweight decoders (or last layers) are trained independently for each frame group, enabling parallelization and efficient scaling to longer videos.}
    \label{Architecture Diagram}
\end{figure}

Here, we will introduce our video encoding pipeline.
SIEDD consists of a two-stage training process. First we train a shared encoder model using a small subset $N_s$ out of $N$ video frames ($N_s \ll N$). In the next stage, we freeze the trained encoder and only train separate decoder networks for each frame group $N_g$. In all our experiments, we held $N_s=N_g$.

\subsection{Shared Encoder Training}

We define the shared encoder as an MLP \( f_\theta: \mathbb{R}^{\text{in}} \rightarrow \mathbb{R}^{d} \), which maps input coordinates to a latent representation. Each \textit{decoder} \( g_{\phi,i}: \mathbb{R}^d \rightarrow \mathbb{R}^3 \), where \( 0 < i < N_s \), is also an MLP that maps the shared latent vector to an RGB output for frame \( i \).
Prior to the encoder, we apply a positional embedding \( \gamma: \mathbb{R}^2 \rightarrow \mathbb{R}^{\text{in}} \) to the 2D input coordinates. 
Both the encoder and decoder use the sine activation function \cite{sitzmann2020implicit} with frequency parameter \( \omega = 30 \) in all layers except the final one, which is left linear to produce the output.
A SIEDD network is composed of a frozen positional embedding, followed by a shared encoder, and finally the $N_s$ decoders. 
This is defined as 
\begin{align*}
    h_{\phi, \theta}(x) &= \text{concat}_i\left( g_{\phi,i} \circ f_\theta \circ \gamma (x) \right) \\
    h &: \mathbb{R}^2 \rightarrow \mathbb{R}^{N_s \times 3}
\end{align*}
We initially fit a shared encoder by overfitting it to $N_s$ separate decoders on uniformly sampled keyframes from the whole video. To achieve this, we optimize 
\begin{align*}
    \theta^*, \phi^* &= \arg\min_{\theta, \phi} \; L(h_{\phi, \theta}(x), y)
\end{align*}
These trained decoders are used to initialize the weights of frame specific decoders in the next stage.
We use the standard $L2$ loss for all of our experiments unless specified otherwise. 

\subsection{Discrete Decoder Training}
We chunk our videos into groups of $N_g$ frames each. We take the trained shared encoder from stage-1 and freeze it, while proceeding to train individual decoders for each frame group. 
To improve the speed, decoder weights are initialized from the closest key frame's decoder weights from the shared encoder model. This method deviates from \citep{vyas2025learningtransferablefeaturesimplicit} where the shared encoder is also trained for unseen images. Note that since our encoder is frozen, these decoders are not dependent on each other, allowing us to train them in parallel, across devices.

\subsubsection{Sharing Decoder Weights}
Using the same architecture as the shared encoder model for the video frame fitting is highly parameter inefficient due to independent weights between similar frames. Therefore, in the second stage of the pipeline, we combine the $N_g$ separate decoder MLPs into a shared decoder for the frame group. However, the last layer of the decoder must remain separate to allow for precise prediction of pixel colors. This approaches vastly improves video compression. We employ BatchLinear layers to speed up matmuls in the decoder, allowing us to decode entire frame groups at once.

\subsubsection{Coordinate Sampling}
Efficient coordinate sampling was critical to achieve low encoding time for SIEDD. A forward pass using a 1080p image's (x, y) coordinates of shape $(1920 \cdot 1080) \times 2$ consumes excessive GPU VRAM and is extremely computationally heavy. While this is necessary to reconstruct the image, we find sampling can greatly speed up training. The $N$ coordinates we use while training are effectively different data samples and it is not necessary to train on each one in every iteration. We use uniform random sampling to sample $C$ points where $C \ll H \dot W$.
To reduce the overhead of random sampling, we shuffle all coordinates once per epoch and iterate through them sequentially, with each minibatch containing $C$ frame coordinates.
We also found an approximate lower limit for C, which was $\approx \frac{H \cdot W}{1024}$ which accelerates training while causing minimal loss to reconstruction quality. For 1080p, this decreases our batch size from 2e6 to 2e3, a $1000 \times$ reduction.

\subsection{Compression Pipeline}

The shared encoder is not quantized due to its insignificant contribution to the overall parameters. The decoders for all video frames undergo post-training quantization. In particular, Half Quadratic Quantization (HQQ) \citep{badri2023hqq} was the optimal method for compressing model weights while retaining reconstruction quality. An important note is that the last layers of the decoders, which produce the output pixels, are kept unquantized to preserve reconstruction quality.
After the model weights are quantized, they are compressed further using huffman encoding. Finally, the resulting bitstream is saved to the disk using lzma-based compression.\footnote{\url{https://github.com/lucianopaz/compress_pickle}}


\section{Experiments}
\subsection{Datasets and Implementation}
We perform experiments using multiple datasets including UVG \citep{10.1145/3339825.3394937}, DAVIS \citep{Perazzi2016}, and Big Buck Bunny. We experimented on 7 UVG-HD videos (Beauty, Bosphorus, HoneyBee, Jockey, ReadySteadyGo, ShakeNDry, and YachtRide) containing a total of 3900 $1920 \times 1080$ video frames. For 4k experiments, we used the same 7 UVG-4k videos with image sizes of $3840 \times 2160$. For the DAVIS dataset, we use 10 1080p videos from the validation set (blackswan, bmx-trees, boat, breakdance, camel, car-roundabout, car-shadow, cows, dance-twirl, and dog). In total, this subset of DAVIS contains 748 frames. Finally, to study long video performance, we use one video from Youtube-8M \citep{abuelhaija2016youtube8mlargescalevideoclassification} about \href{https://www.youtube.com/watch?v=4yZlK2Ftjho}{Mario Kart}. The video was downloaded at $1280 \times 720$ resolution and the first 4000 frames were used. 
We use ffmpeg to convert the raw YUV files into PNG frames. Further details are included  in the Appendix. 

We use the standard metrics of PSNR (Peak signal to noise ratio) and SSIM (Structural similarity) as our primary measures of video quality. We use BPP (bits per pixel) to measure the compression efficiency. 
All encoding time measures for all models are for a single NVIDIA RTXA5000 GPU. Additional quality and speed metrics are included in the supplementary.


\begin{figure*}[t]
  \centering
  \begin{subfigure}[b]{0.49\linewidth}
    \centering
    \includegraphics[width=\linewidth]{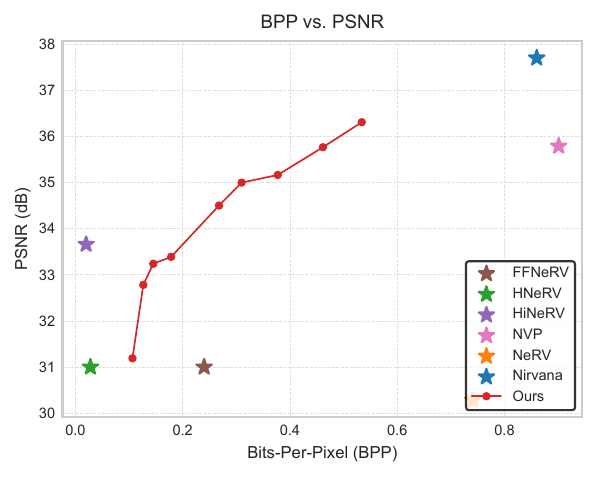}
    \caption{BPP vs.\ PSNR on UVG-HD}
    \label{fig:bpp-psnr}
  \end{subfigure}
  \begin{subfigure}[b]{0.49\linewidth}
    \centering
    \includegraphics[width=\linewidth]{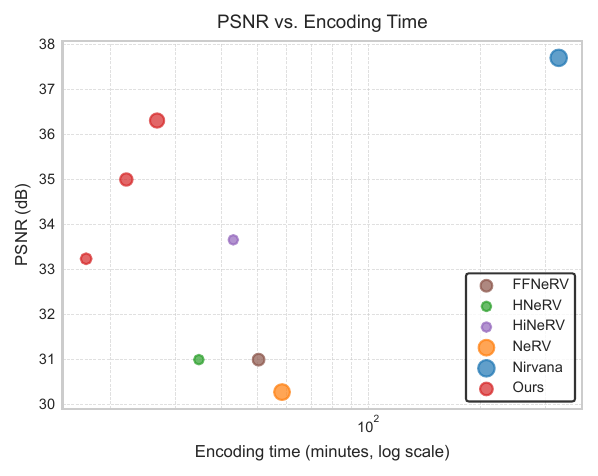}
    \caption{PSNR vs.\ encoding time (log-minutes).}
    \label{fig:psnr-time}
  \end{subfigure}
  %
  \caption{Comparison of our method and baselines on UVG.  
           Left: rate–distortion; Right: speed–quality trade-off.}
  \label{fig:uvg-plots}
\end{figure*}

\subsection{Setup}

Our models were implemented using PyTorch and experiments were run using NVIDIA RTX A5000 GPUs. We used the schedule-free AdamW optimizer \citep{defazio2024roadscheduled} which provided stable training when compared to commonly used optimizers such as those of the Adam family. We define 3 SIEDD models: SIEDD-S with a model dimension of 512, SIEDD-M with a model dimension of 768, and SIEDD-L with a model dimension of 1024. For all 3 models, the shared encoder has 1 hidden layer while the decoders contain 3. The number of iterations for the shared encoder training and the video frame training were kept equal at 20000.

\subsection{HD Video Reconstruction}
We showcase SIEDD’s ability to efficiently encode 1080p videos from the UVG-HD dataset. In Figure~\ref{fig:bpp-psnr}, we compare the rate-distortion trade-off of our method against several frame-based Video-INR baselines \cite{chen2021nerv,chen2023hnerv,kwan2023hinerv,lee2023ffnerv,maiya2023nirvana}, all given a maximum encoding time budget of 1 hour for a single 600-frame, 1080p clip. SIEDD achieves high-quality reconstructions at competitive compression levels, outperforming several methods that either sacrifice quality (e.g., HiNeRV\cite{kwan2023hinerv}) or require substantially more bandwidth (e.g., Nirvana \cite{maiya2023nirvana}, FFNeRV\cite{lee2023ffnerv}).

Figure~\ref{fig:psnr-time} visualizes the trade-off between PSNR and encoding time (log-scale), with marker size proportional to BPP. SIEDD consistently achieves the best quality-per-time ratio. Notably, it is approximately 20× faster than the closest high-quality baseline while operating at lower or comparable bitrates. Unlike FFNeRV \cite{lee2023ffnerv} or NIRVANA \cite{maiya2023nirvana} which cluster in the high-time, high-rate regime, SIEDD pushes the pareto front forward—offering both efficiency and fidelity.

A key observation is that some methods like NeRV\cite{chen2021nerv} and HNeRV\cite{chen2023hnerv} attain decent reconstruction quality but at significantly higher encoding costs, making them less practical. SIEDD's performance illustrates the advantage of its shared encoder design, efficient sampling strategy, and decoder parallelizability. This positions SIEDD not only as a competitive Video-INR in terms of compression, but as a viable candidate for real-world, time-sensitive deployment scenarios.

\begin{figure}[t]
    \centering
    \begin{minipage}{0.45\textwidth}
        \begin{tikzpicture}
            \node[inner sep=0pt] (img) at (0,0) {\includegraphics[width=\linewidth]{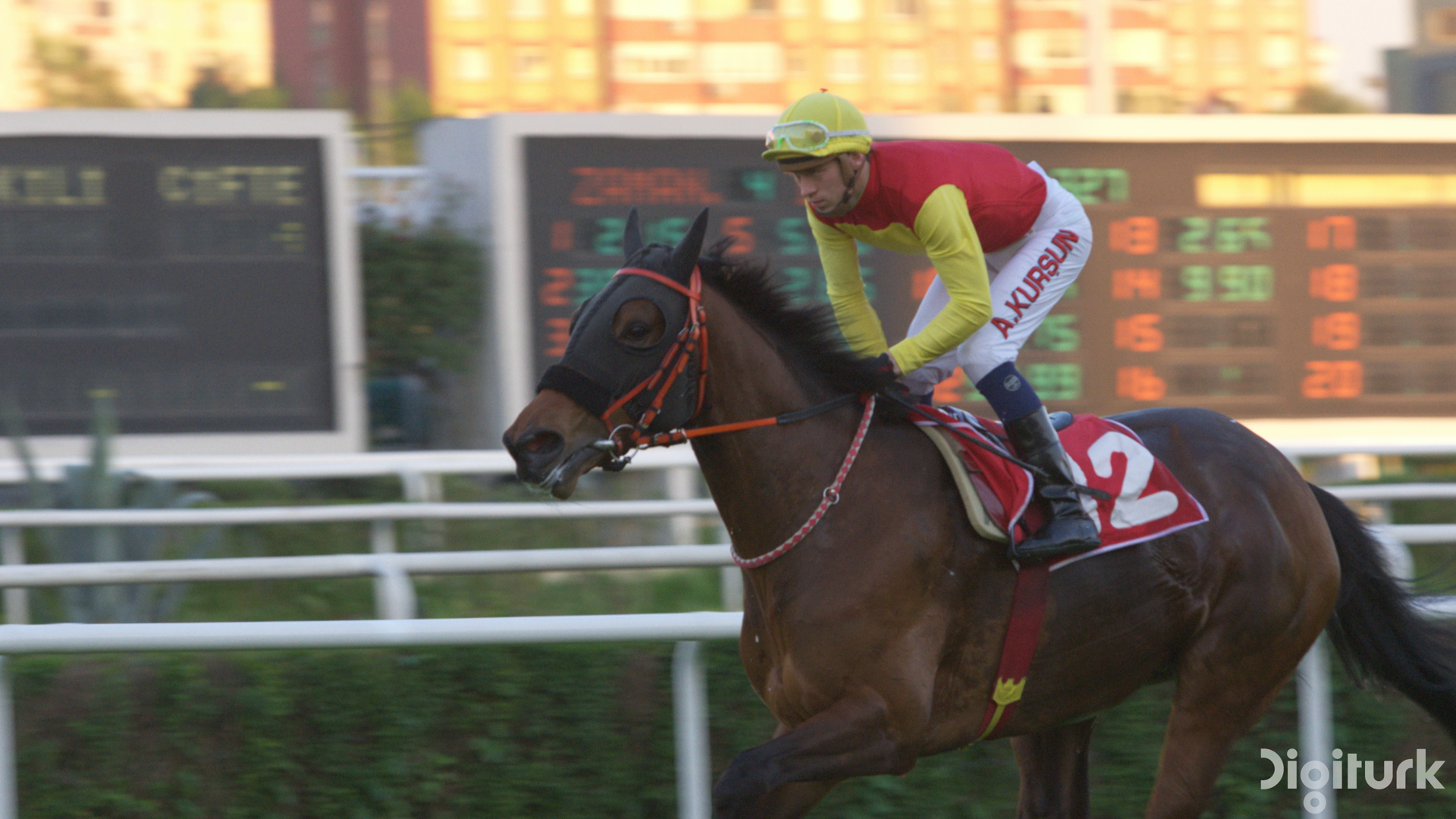}};
            \draw[red, thick] (-0.8,1.4) rectangle (0.8,0.3);
            \node[inner sep=0pt] (img2) at (1.64, -0.92)
            {\includegraphics[width=0.48\linewidth]{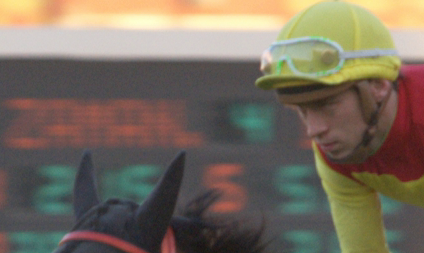}};
            \node at (2.5,-1.6) [fill=white, text=black] {\tiny Ground Truth};
        \end{tikzpicture}
    \end{minipage}
    \begin{minipage}{0.45\textwidth}
        \begin{tikzpicture}
            \node[inner sep=0pt] (img) at (0,0) {\includegraphics[width=0.48\linewidth]{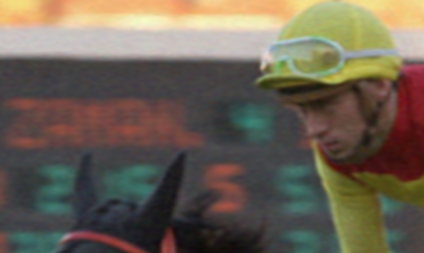}};
            \node[anchor=south east] at (img.south east) [fill=white, inner sep=1pt] {\tiny SIEDD (Ours)};
        \end{tikzpicture}%
        \begin{tikzpicture}
            \node[inner sep=0pt] (img) at (0,0) {\includegraphics[width=0.48\linewidth]{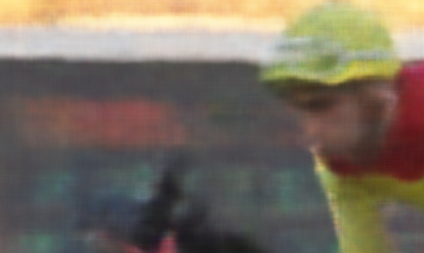}};
            \node[anchor=south east] at (img.south east) [fill=white, inner sep=1pt] {\tiny HNeRV};
        \end{tikzpicture}\\[0.3em]
        \begin{tikzpicture}
            \node[inner sep=0pt] (img) at (0,0) {\includegraphics[width=0.48\linewidth]{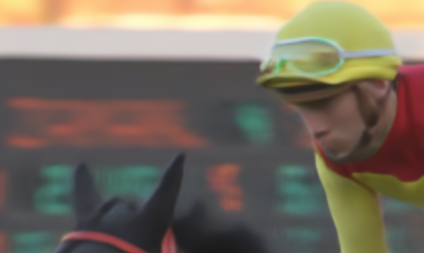}};
            \node[anchor=south east] at (img.south east) [fill=white, inner sep=1pt] {\tiny HiNerv};
        \end{tikzpicture}%
        \begin{tikzpicture}
            \node[inner sep=0pt] (img) at (0,0) {\includegraphics[width=0.48\linewidth]{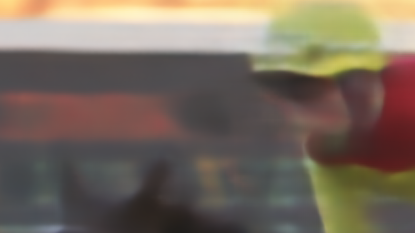}};
            \node[anchor=south east] at (img.south east) [fill=white, inner sep=1pt] {\tiny FFNeRV};
        \end{tikzpicture}
    \end{minipage}
    \begin{minipage}{0.45\textwidth}
        \begin{tikzpicture}
            \node[inner sep=0pt] (img) at (0,0) {\includegraphics[width=\linewidth]{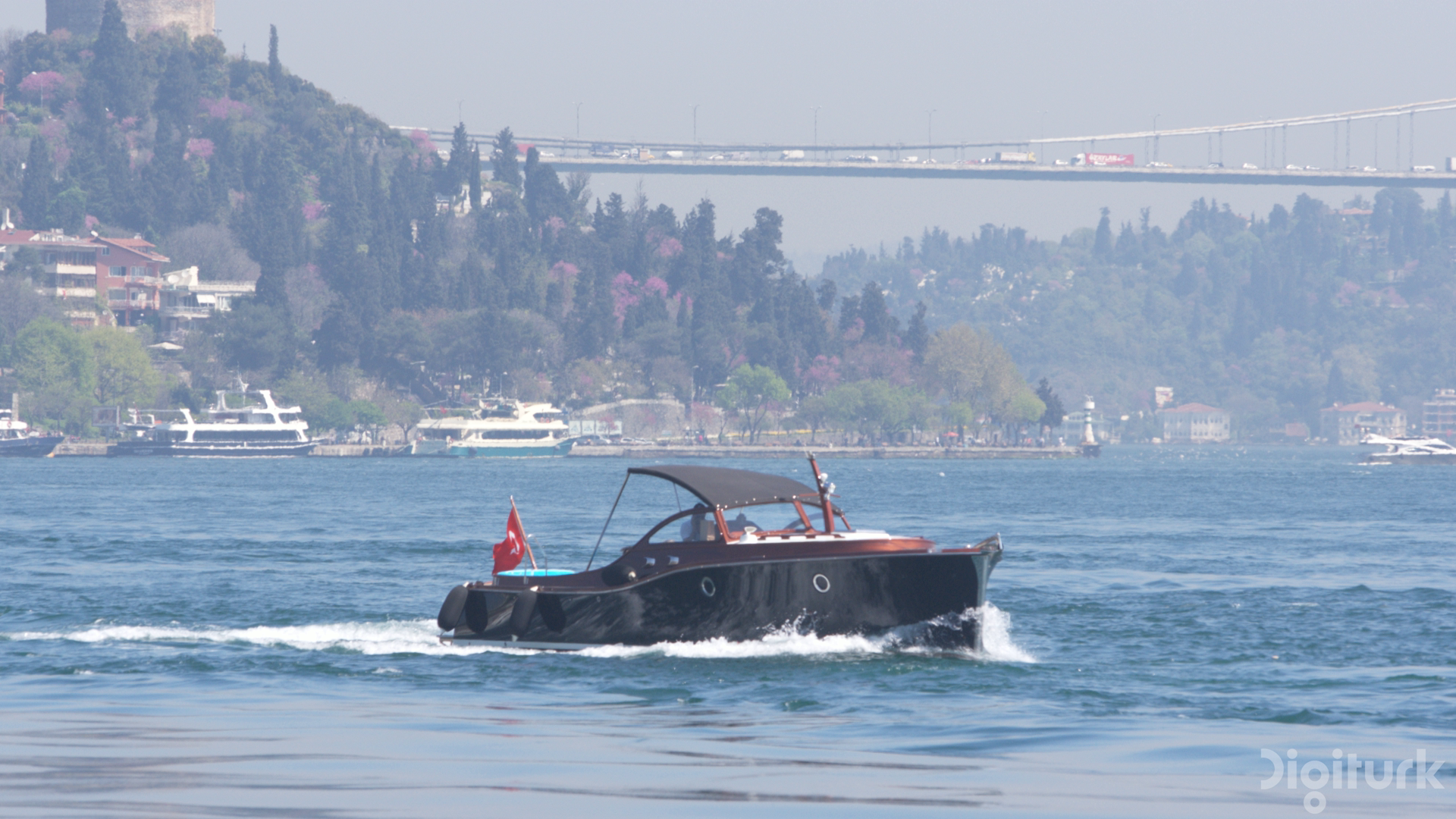}};
            \draw[red, thick] (-1.3,-0.4) rectangle (-0.6,-0.8);
            \node[inner sep=0pt] (img2) at (1.64, -0.92)
            {\includegraphics[width=0.48\linewidth]{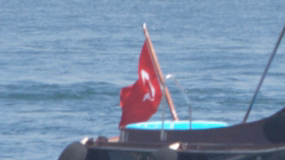}};
            \node at (2.5,-1.6) [fill=white, text=black] {\tiny Ground Truth};
        \end{tikzpicture}
    \end{minipage}
    \begin{minipage}{0.45\textwidth}
        \begin{tikzpicture}
            \node[inner sep=0pt] (img) at (0,0) {\includegraphics[width=0.48\linewidth]{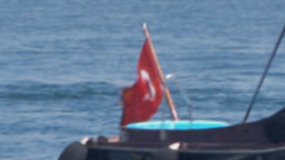}};
            \node[anchor=south east] at (img.south east) [fill=white, inner sep=1pt] {\tiny SIEDD (Ours)};
        \end{tikzpicture}%
        \begin{tikzpicture}
            \node[inner sep=0pt] (img) at (0,0) {\includegraphics[width=0.48\linewidth]{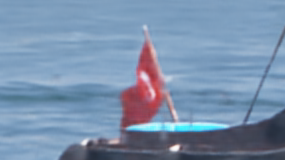}};
            \node[anchor=south east] at (img.south east) [fill=white, inner sep=1pt] {\tiny HNeRV};
        \end{tikzpicture}\\[0.3em]
        \begin{tikzpicture}
            \node[inner sep=0pt] (img) at (0,0) {\includegraphics[width=0.48\linewidth]{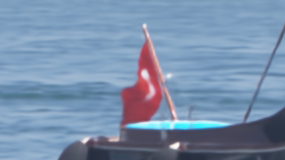}};
            \node[anchor=south east] at (img.south east) [fill=white, inner sep=1pt] {\tiny HiNerv};
        \end{tikzpicture}%
        \begin{tikzpicture}
            \node[inner sep=0pt] (img) at (0,0) {\includegraphics[width=0.48\linewidth]{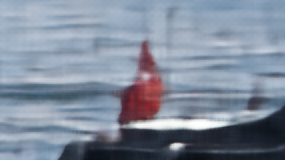}};
            \node[anchor=south east] at (img.south east) [fill=white, inner sep=1pt] {\tiny FFNeRV};
        \end{tikzpicture}
    \end{minipage}
    \caption{Video Reconstruction Visualization. We compare the reconstructions of SEIDD with other baselines for 2 UVG-HD Videos: Jockey (top) and Bosphorus (Bottom). We can clearly see that SEIDD produces much sharper reconstructions, staying true to the ground truth.}
    \label{fig:recon-viz}
\end{figure}
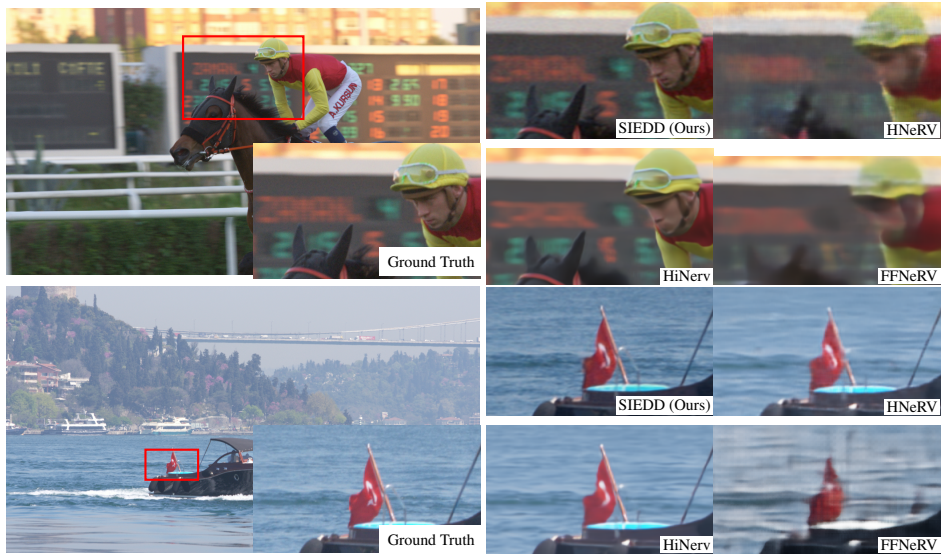

In Fig \ref{fig:recon-viz} we visualize the reconstructed frames from SEIDD and other baselines for ``jockey" and the ``bosphorous" sequence from UVG-HD \cite{10.1145/3339825.3394937} dataset. We can clearly see that our method preserves high frequency details and does not have the \textit{smudging/smoothing} effect that is visible in the baselines. 

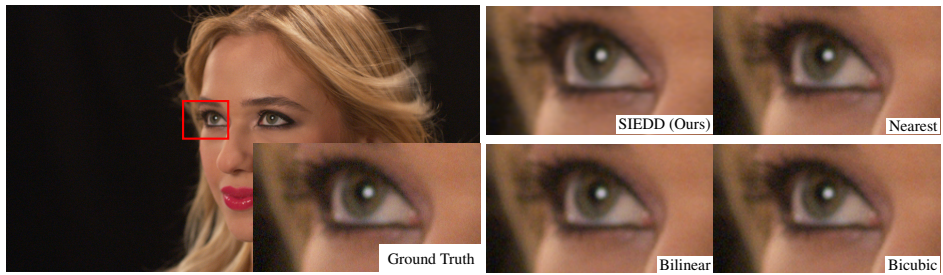
\begin{figure}[t]
    \centering

    \begin{minipage}{0.45\textwidth}
        \begin{tikzpicture}
            \node[inner sep=0pt] (img) at (0,0) {\includegraphics[width=\linewidth]{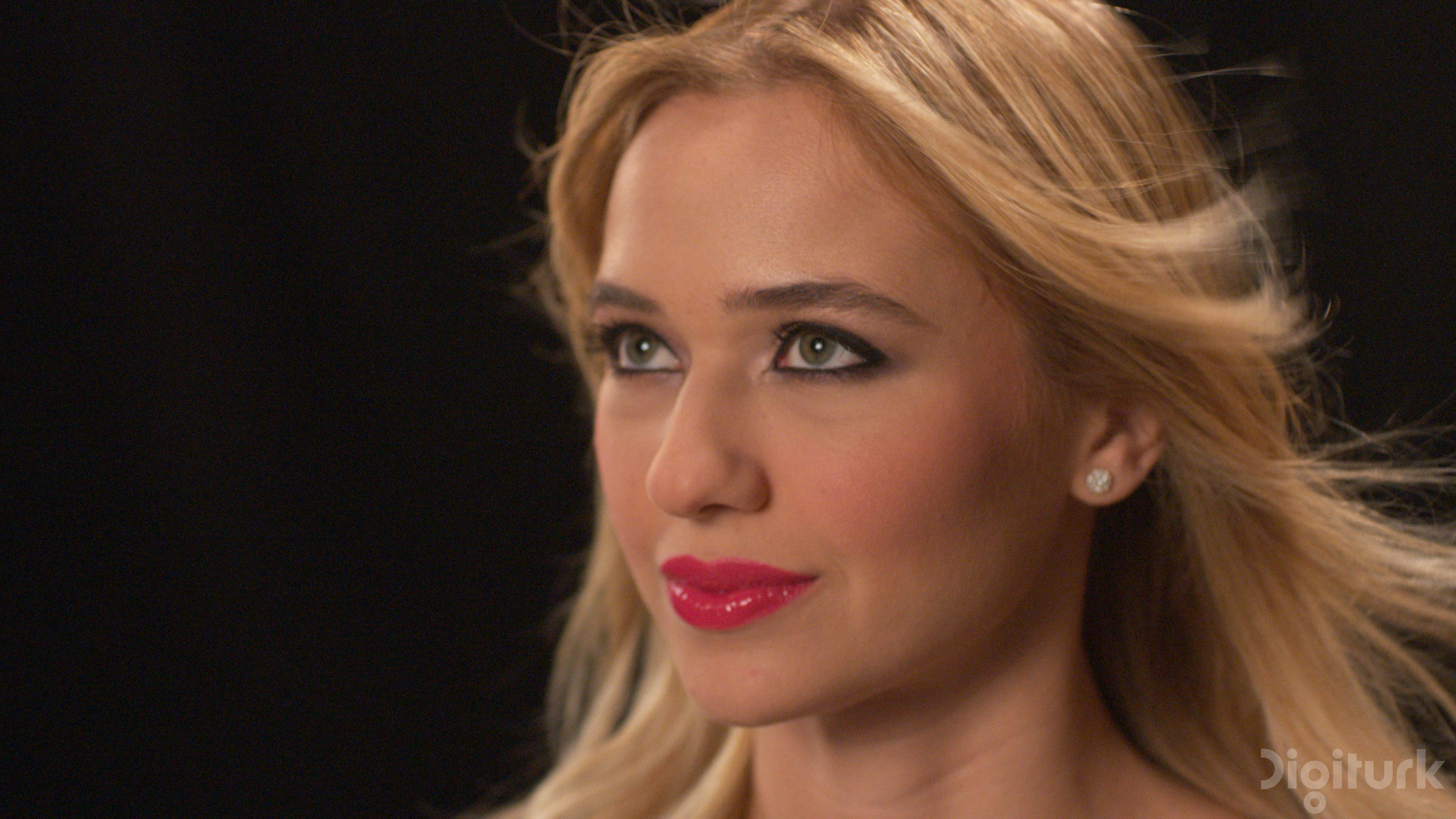}};
            \draw[red, thick] (-0.8,0.5) rectangle (-0.2,0);
            \node[inner sep=0pt] (img2) at (1.64, -0.92)
            {\includegraphics[width=0.48\linewidth]{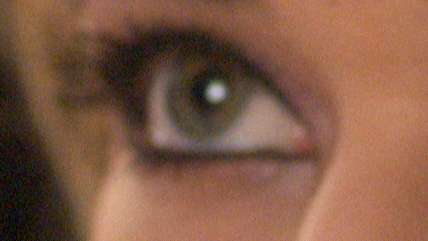}};
            \node at (2.5,-1.6) [fill=white, text=black] {\tiny Ground Truth};
        \end{tikzpicture}
    \end{minipage}
    \begin{minipage}{0.45\textwidth}
        \begin{tikzpicture}
            \node[inner sep=0pt] (img) at (0,0) {\includegraphics[width=0.48\linewidth]{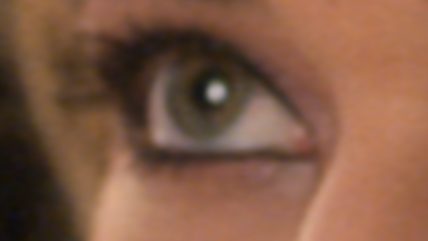}};
            \node[anchor=south east] at (img.south east) [fill=white, inner sep=1pt] {\tiny SIEDD (Ours)};
        \end{tikzpicture}%
        \begin{tikzpicture}
            \node[inner sep=0pt] (img) at (0,0) {\includegraphics[width=0.48\linewidth]{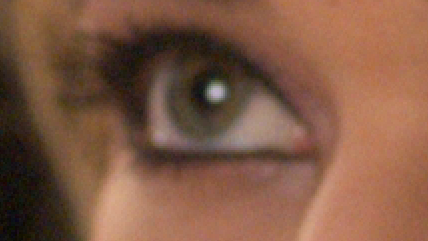}};
            \node[anchor=south east] at (img.south east) [fill=white, inner sep=1pt] {\tiny Nearest};
        \end{tikzpicture}\\[0.3em]
        \begin{tikzpicture}
            \node[inner sep=0pt] (img) at (0,0) {\includegraphics[width=0.48\linewidth]{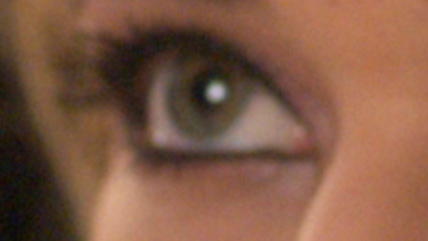}};
            \node[anchor=south east] at (img.south east) [fill=white, inner sep=1pt] {\tiny Bilinear};
        \end{tikzpicture}%
        \begin{tikzpicture}
            \node[inner sep=0pt] (img) at (0,0) {\includegraphics[width=0.48\linewidth]{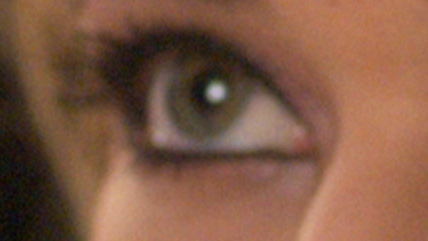}};
            \node[anchor=south east] at (img.south east) [fill=white, inner sep=1pt] {\tiny Bicubic};
        \end{tikzpicture}
    \end{minipage}
    \caption{Super resolution visualization between baseline methods (nearest, bilinear, bicubic) and SIEDD-L. Upon close inspection, it is visible that the noise present in the ground truth image is not represented by SIEDD compared to other methods.}
\end{figure}

\subsection{4K Video Reconstruction}

\begin{table}
    \begin{minipage}[t]{0.45\textwidth}
        \centering
        \caption{Shared Encoder Weight Transfer from UVG-HD to DAVIS}
        \begin{tabular}{@{}lccc@{}}
            \toprule
            Method & PSNR $\uparrow$ & SSIM $\uparrow$ & Time (s) $\downarrow$ \\
            \midrule
            SIEDD-M & 30.71 & 0.82 & 442 \\
            From UVG-HD & 30.63 & 0.83 & 158\\
            \bottomrule
        \end{tabular}
        \label{tab:DAVIS}
    \end{minipage}
    \begin{minipage}[t]{0.6\textwidth}
        \centering
        \caption{Long Video Results}
         \begin{tabular}{@{}lccc@{}}
        \toprule
        Method & PSNR $\uparrow$ & BPP $\downarrow$ & Time (s) $\downarrow$ \\
        \midrule
        Ours & 31.10 & 0.325 & 6408 \\
        HNeRV & 24.78 & 0.035 & 7800 \\
        NeRV & 23.94 & 0.12 & 11000 \\
        HiNeRV & 26.68 & 0.04 & 6860 \\
        \bottomrule
    \end{tabular}        
            
        \label{tab:long}
    \end{minipage}
\end{table}
\begin{table}[t]
    \centering
    \caption{4K reconstruction results on UVG-4K. SIEDD variants compared with NeRV, HiNeRV, and Nirvana. Encoding Time reported in seconds.}
    \begin{tabular}{@{}lccccc@{}}
        \toprule
        Method & PSNR $\uparrow$ & SSIM $\uparrow$ & BPP $\downarrow$ & FPS $\uparrow$ & Encoding Time (s) $\downarrow$ \\
        \midrule
        SIEDD-M & 33.41 & 0.83 & 0.078 & 1.16 & 2574 \\
        SIEDD-M $N_g=10$ & 34.42 & 0.84 & 0.147 & 1.14 & 2223 \\
        SIEDD-L, $N_g=10$ & 35.41 & 0.85 & 0.256 & 0.76 & 3033 \\
        \midrule
        $3{\times}3$ Patches & 34.52 & 0.84 & 0.275 & 7.49 & 3160 \\
        $6{\times}6$ Patches & 33.70 & 0.82 & 0.331 & 21.70 & 4211 \\
        \midrule
        NeRV & 28.67 & 0.83 & 0.486 & 6.3 & 3660 \\
        HiNeRV & 30.47 & 0.86 & 0.0051 & 21.0 & 10200 \\
        Nirvana & 35.18 & 0.93 & 0.270 & 28.4 & 75600 \\
        \bottomrule
    \end{tabular}
    \label{tab:UVG-4k}
\end{table}

We additionally show SIEDD's ability to encode 4k video and present the results for 3 different model configurations in \ref{tab:UVG-4k}. We can see that SEIDD clearly outperforms all the baselines in terms of reconstruction quality with excellent encoding speeds. In fact, this is the first Video-INR method that is able to encode a 600-frame 4K video in under an hour, achieving upto 30X faster times compared to baselines.
We additionally test two SIEDD-L models on UVG-4k with $3 \times 3$ and $6 \times 6$ patches. We use square patches of size $p \times p$, meaning that for every image coordinate, SIEDD will output the colors of $p^2$ pixels. This reduces the total number of image coordinates by a factor of $p^2$ while increasing the number of parameters in the last layer of the decoder by $p^2$ as a tradeoff. As shown in \ref{tab:UVG-4k}, decoding speed drastically improves with associated trade-offs in encoding time, and compression.

\subsubsection{Shared Encoder Transfer}

We also show the possibility of shared encoder transfer across datasets in \ref{tab:DAVIS}. To show this, we use the weights of the shared encoder trained on UVG-HD and use it to train DAVIS in place of the shared encoder training process. We compare this to training SEIDD on DAVIS from scratch and 
find that using a shared encoder from UVG-HD provides similar reconstruction quality while reducing the encoding time significantly due to the lack of shared encoder training. This points us towards a broader idea - that the shared encoder is like an ever growing ``prior" of videos which can be updated when required, else used to perform zero-shot transfer to unseen videos.

\begin{figure*}[t]
  \centering
  \begin{subfigure}[b]{0.49\linewidth}
    \centering
    \includegraphics[width=\linewidth]{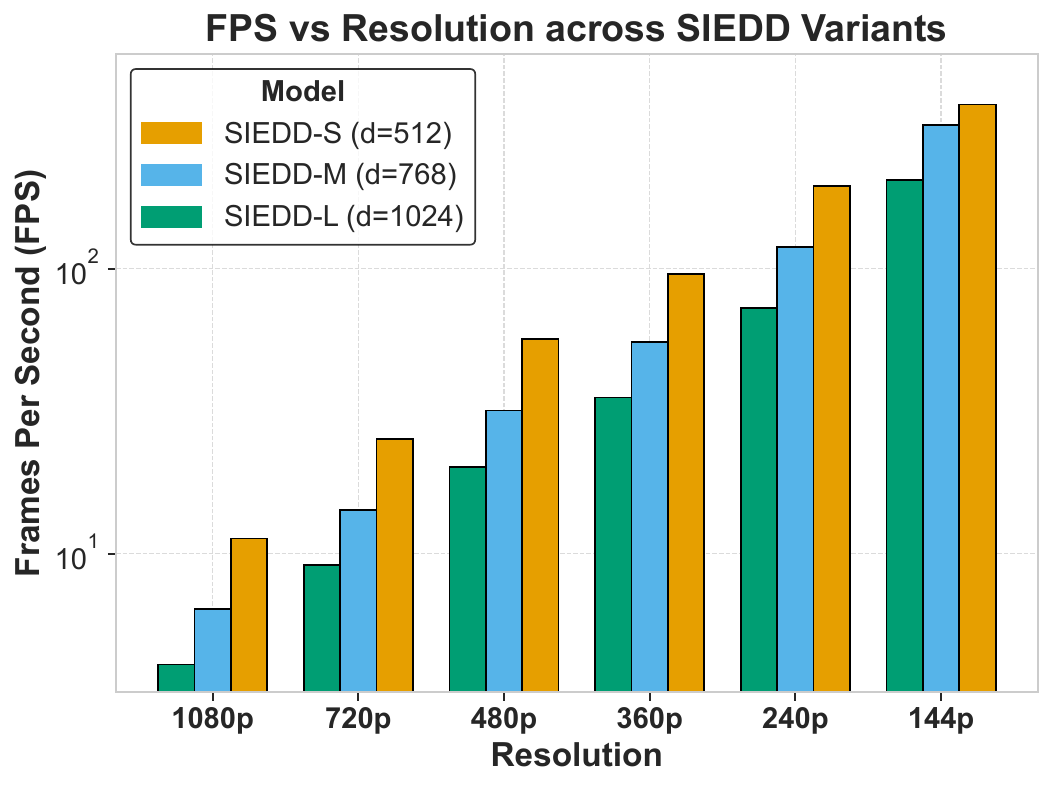}
    \caption{FPS vs. Resolution on UVG-HD.}
    \label{fig:any-fps}
  \end{subfigure}
  \begin{subfigure}[b]{0.49\linewidth}
    \centering
    \includegraphics[width=\linewidth]{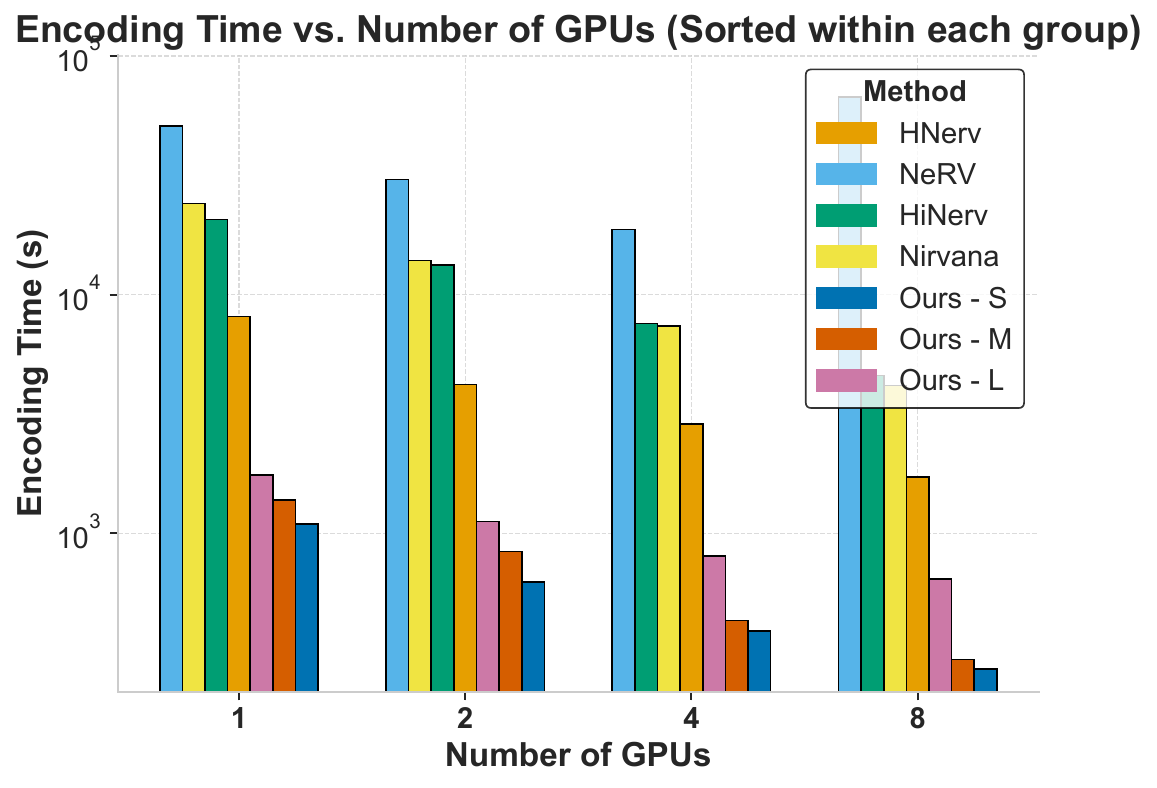}
    \caption{Encoding time and GPU Parallelization for SIEDD and baselines.}
    \label{fig:gpu-parallel}
  \end{subfigure}
  %
  \caption{Comparison of our method and baselines on UVG.  
           Left: rate–distortion; Right: speed–quality trade-off.}
  \label{fig:speed}
\end{figure*}

\subsection{SuperResolution}
We demonstrate SIEDD's capability for continuous-resolution decoding by comparing its super-resolved outputs against traditional interpolation methods. As shown in Figure~4, SIEDD-L reconstructs fine details such as eyelashes and iris contours with greater fidelity compared to nearest, bilinear, and bicubic upsampling. Interestingly, the noise texture seen in the ground truth is absent in SIEDD's reconstruction, suggesting that the model implicitly denoises while super-resolving. Unlike baselines that rely on fixed grid interpolation, SIEDD leverages learned implicit mappings to reconstruct semantically meaningful detail, making it well-suited for applications requiring resolution-adaptive decoding.

\subsection{Any Resolution Decoding}
Unlike Frame-based Video-INRs our method takes 2D positional grid as input which allows us to control the spatial resolution of the decoded output. In Figure~\ref{fig:any-fps} we measure how the decoding speed is impacted at different resolutions and observe a consistent pattern of faster decoding at lower resolutions. 

\subsection{GPU parallelization}
We evaluate how encoding time scales with the number of GPUs for various Video-INR baselines and our SIEDD variants. As shown in Figure~\ref{fig:gpu-parallel}, SIEDD exhibits near-linear scaling with increasing GPU count. Specifically, our largest model (SIEDD-L) shows a consistent drop in encoding time from 1 GPU to 8 GPUs, highlighting the effectiveness of parallel decoder training across frame groups.

Compared to baselines like NeRV~\cite{chen2021nerv} and HiNeRV~\cite{kwan2023hinerv}, which show limited gains with more GPUs due to their sequential or monolithic training structure, SIEDD benefits directly from architectural parallelism. For example, at 8 GPUs, SIEDD-L achieves a ~8× speedup relative to its single-GPU decoder training time, while NeRV improves by only ~4×. Even smaller SIEDD variants outperform stronger baselines like Nirvana, despite using fewer parameters and lower computational overhead.

This scalability makes SIEDD a compelling choice for high-resolution or long video scenarios where encoding throughput is critical.

\subsection{Long Video Training}
To test the scaling of our model on the temporal axis, we train on a sequence of 4000 frames from ``mario-kart'' sequence from Youtube-8M dataset. The results are presented in Table \ref{tab:long} and we see that SEIDD outperforms other baselines with great encoding speeds.

\begin{table}[t]
    \centering
        \caption{Ablation studies: (a) Effect of coordinate sampling rate on reconstruction quality and encoding time. (b) Effect of shared encoder training iterations on reconstruction quality.}
            \label{tab:ablations}
    \begin{minipage}{0.48\textwidth}
        \centering
                \caption*{\textbf{(a)} Sampling rate vs. quality and encoding time.}
        \begin{tabular}{@{}lccc@{}}
            \toprule
            Sampling Rate & PSNR (dB)$\uparrow$&SSIM $\uparrow$&Time (s)$\downarrow$ \\
            \midrule
            1/128  & 33.30 & 0.766 & 5318.51 \\
            1/256  & 33.28 & 0.765 & 2790.15 \\
            1/512  & 33.27 & 0.765 & 1556.00 \\
            1/1024 & 33.24 & 0.765 & 887.49  \\
            1/2048 & 32.72 & 0.758 & 618.18  \\
            \bottomrule
        \end{tabular}
        \label{tab:sampling-rate-ablation}
    \end{minipage}
    \hfill
    \begin{minipage}{0.48\textwidth}
        \centering
                \caption*{\textbf{(b)} Shared encoder iterations vs. quality.}
        \begin{tabular}{@{}lcc@{}}
            \toprule
            Shared Iters & PSNR (dB) $\uparrow$ & SSIM $\uparrow$ \\
            \midrule
            500   & 35.15 & 0.886 \\
            2000  & 35.23 & 0.886 \\
            5000  & 35.27 & 0.886 \\
            \bottomrule
        \end{tabular}
    \end{minipage}
\end{table}

\subsection{Ablation Analysis}
\subsubsection{Sampling Rate}
We perform an ablation study on the coordinate sampling rate to understand its impact on encoding time and reconstruction quality. As shown in Table ~\ref{tab:ablations}(a), reducing the sampling rate from 1/128 to 1/2048 leads to a significant drop in encoding time—from 5318s to just 618s—demonstrating a nearly 9× speedup. Notably, the reconstruction quality (PSNR and SSIM) remains largely stable for moderate reductions (up to 1/1024), with only a minor degradation (~0.05 dB PSNR). Beyond this, quality drops more noticeably, indicating diminishing returns. This trade-off highlights the effectiveness of our sparse coordinate sampling strategy, allowing users to balance compute budget and fidelity depending on application needs.

\subsubsection{Shared Encoder iterations}
We study the effect of shared encoder training iterations on reconstruction quality in Table~\ref{tab:ablations}(b). As expected, increasing the number of training steps improves PSNR marginally—from 35.15 at 500 iterations to 35.27 at 5000—while SSIM remains largely unchanged. This suggests that the encoder converges quickly to useful low-frequency features, validating our design choice to limit encoder training for faster overall encoding.

\begin{figure*}[t]
  \centering
  \begin{subfigure}[b]{0.49\linewidth}
    \centering
    \includegraphics[width=\linewidth]{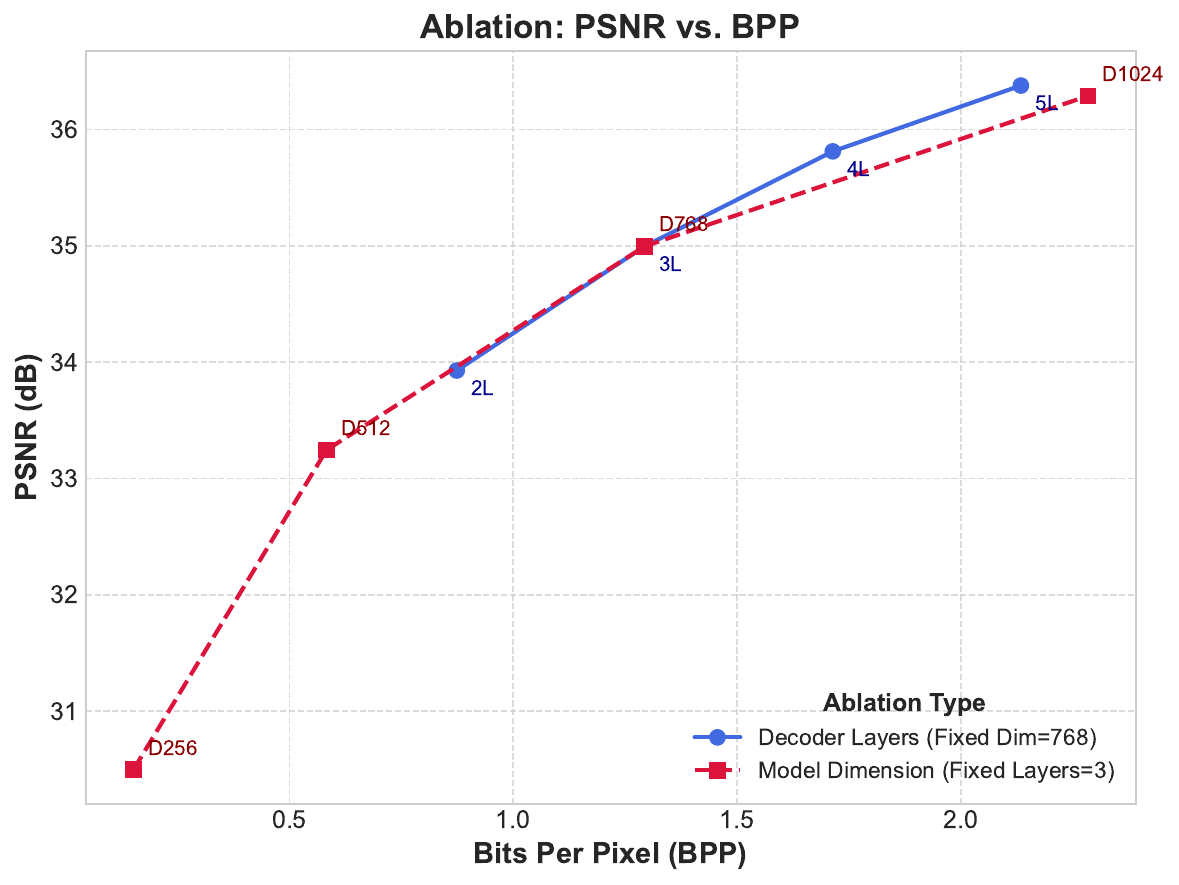}
    \caption{BPP vs.\ PSNR varying model dim and layers}
    \label{fig:dim-depth-ablation-bpp}
  \end{subfigure}
  \begin{subfigure}[b]{0.49\linewidth}
    \centering
    \includegraphics[width=\linewidth]{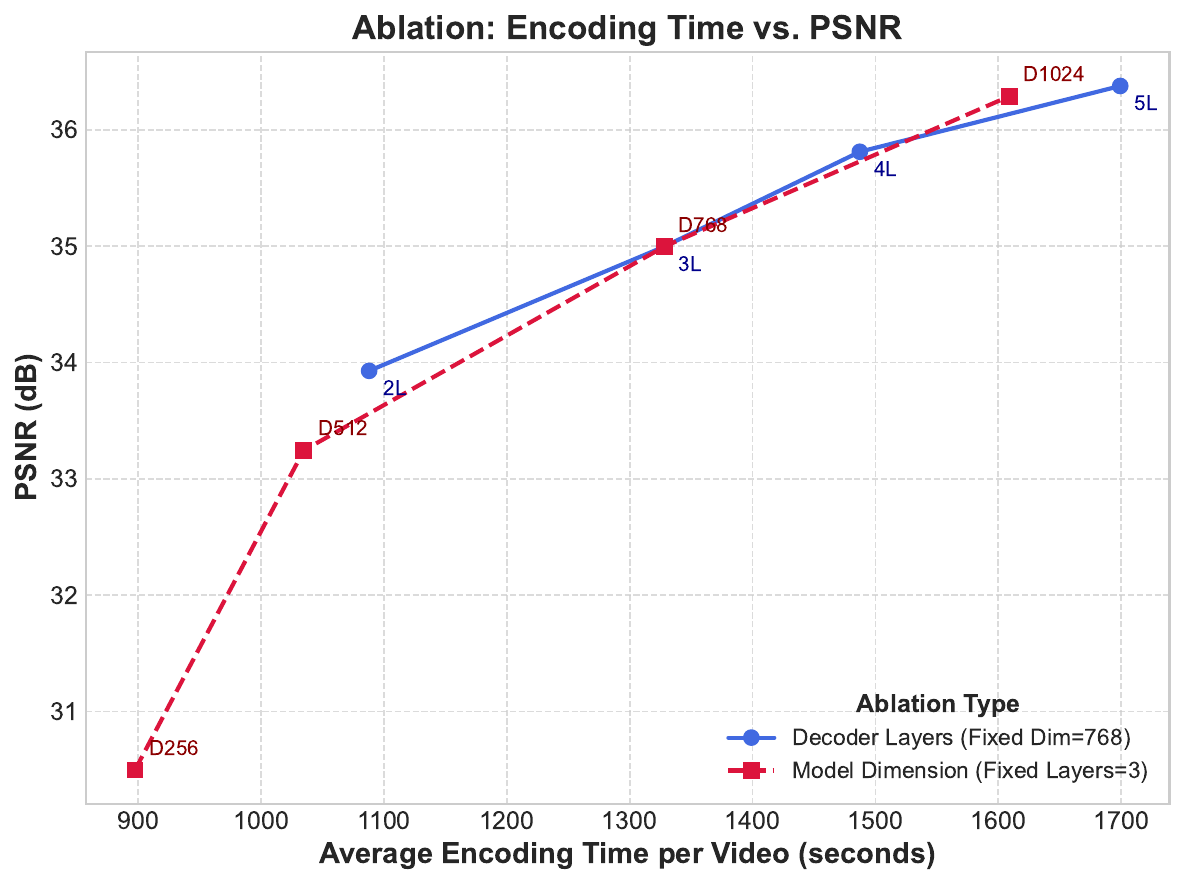}
    \caption{Varying model dim and layers with encoding time}
    \label{fig:dim-depth-ablation-time}
  \end{subfigure}
  %
  \caption{ Ablation study on decoder architecture. (a) PSNR vs.\ BPP when varying decoder layer count (with fixed dim = 768) and model dimension (with fixed 3 decoder layers). (b) PSNR vs.\ encoding time, illustrating trade-offs in reconstruction quality with increased decoder depth or width. More layers improve quality marginally with negligible cost, while increasing model dimension significantly boosts quality at the expense of higher encoding time.}
  \label{fig:uvg-ablation}
\end{figure*}

\subsubsection{Model Layers and Layer dimension}
We investigate how architectural capacity—specifically the number of decoder layers and model dimensionality—affects the rate-distortion performance and encoding time of SIEDD. The results are summarized in Figure~\ref{fig:uvg-ablation}.
In Figure~\ref{fig:dim-depth-ablation-bpp}, we vary the decoder MLP dimension (256, 512, 768, 1024) and also vary the number of decoder layers (2, 3, 4, 5). We observe a consistent improvement in PSNR as dimensionality increases, along with a mild rise in BPP. Notably, the 1024d variant achieves the highest reconstruction quality while maintaining competitive compression, demonstrating that increased latent capacity allows the decoders to model finer visual details more effectively.
In contrast, Figure~\ref{fig:dim-depth-ablation-time} explores the impact of model hyperparameters on encoding speed. While larger decoders yield marginal gains in PSNR, the returns diminish beyond 3 layers and d=768. More importantly, larger networks incur significant increases in encoding time due to slower convergence and additional compute.
Overall, we find that increasing width (dimensionality) provides better quality–bitrate trade-offs, while deeper networks primarily affect training time. These trends guided our default configuration (768d, 3-layer decoder), striking a balance between efficiency and quality.

\section{Conclusion}
We present SIEDD, a fast and scalable Video-INR architecture that leverages shared representations and per-group decoders to dramatically reduce encoding time. By training the encoder on sparse anchor frames and freezing it for the rest of the video, SIEDD enables parallelized decoding and efficient representation learning—achieving up to 30× faster encoding compared to existing INR methods. Our coordinate-based formulation allows continuous-resolution decoding, and our use of simple MLPs makes the architecture amenable to post-training quantization using state-of-the-art compression techniques like HQQ and BNB.

While SIEDD significantly advances the practicality of INR-based codecs, a few areas remain open. Inference-time decoding, especially for high-resolution and high-framerate video, could be further accelerated with fused matmul kernels and specialized hardware-aware optimizations. Our current quantization is post-training; future work could explore quantization-aware training (QAT) to further improve compression without sacrificing fidelity. Finally, while our shared encoder shows strong transfer potential, systematic studies on cross-video generalization and zero-shot inference remain a rich direction for exploration.


\newpage
\appendix
\begin{center}
\Large{\textbf{SIEDD: Shared‑Implicit Encoder with Discrete Decoders}}

\large{Supplementary Material}
\end{center}
\section{Experimental Baseline Settings}
Note that all the following models were trained with a hard limit of 60 minutes on encoding time, on an RTXA5000. We chose to train from scratch as the learning rate schedule has a significant effect on final quality.

\subsection{NeRV}
We use the NeRV-L setting from the original paper~\cite{chen2021nerv}. This comes with 5-NeRV blocks with upscale factors of $[5,3,2,2,2]$ for UVG-HD and $[5,3,2,2,2]$ for UVG-4K. We use the standard hyperparameter settings from the paper. 

\subsection{HiNeRV}
Since HiNeRV \cite{kwan2023hinerv} training can be quite slow, we choose to use the HiNeRV-S configuration from the original paper and add an additional block to make it work for 4K. 

\subsection{FFNeRV}
To balance quality and encoding time, we choose the FFNeRV configuration with $C_{1}, C_{2},S$ as $(112,896,54)$

\subsection{HNeRV} 
We use the default model configuration for UVG-HD as reported in the paper and added an additional block for UVG-4K. Due to architectural constraints, HNeRV ~\cite{chen2023hnerv} output is restricted to $960 \times 1920$ ($~12\%$ less) for UVG-HD and $3600 \times 2160$ for UVG-4K ($~7\%$ less). 

\section{Additional Qualitative Results}

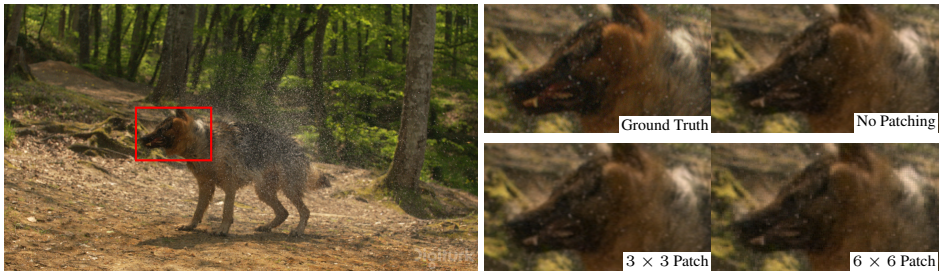
\begin{figure}[b]
    \centering

    \begin{minipage}{0.45\textwidth}
        \begin{tikzpicture}
            \node[inner sep=0pt] (img) at (0,0) {\includegraphics[width=\linewidth]{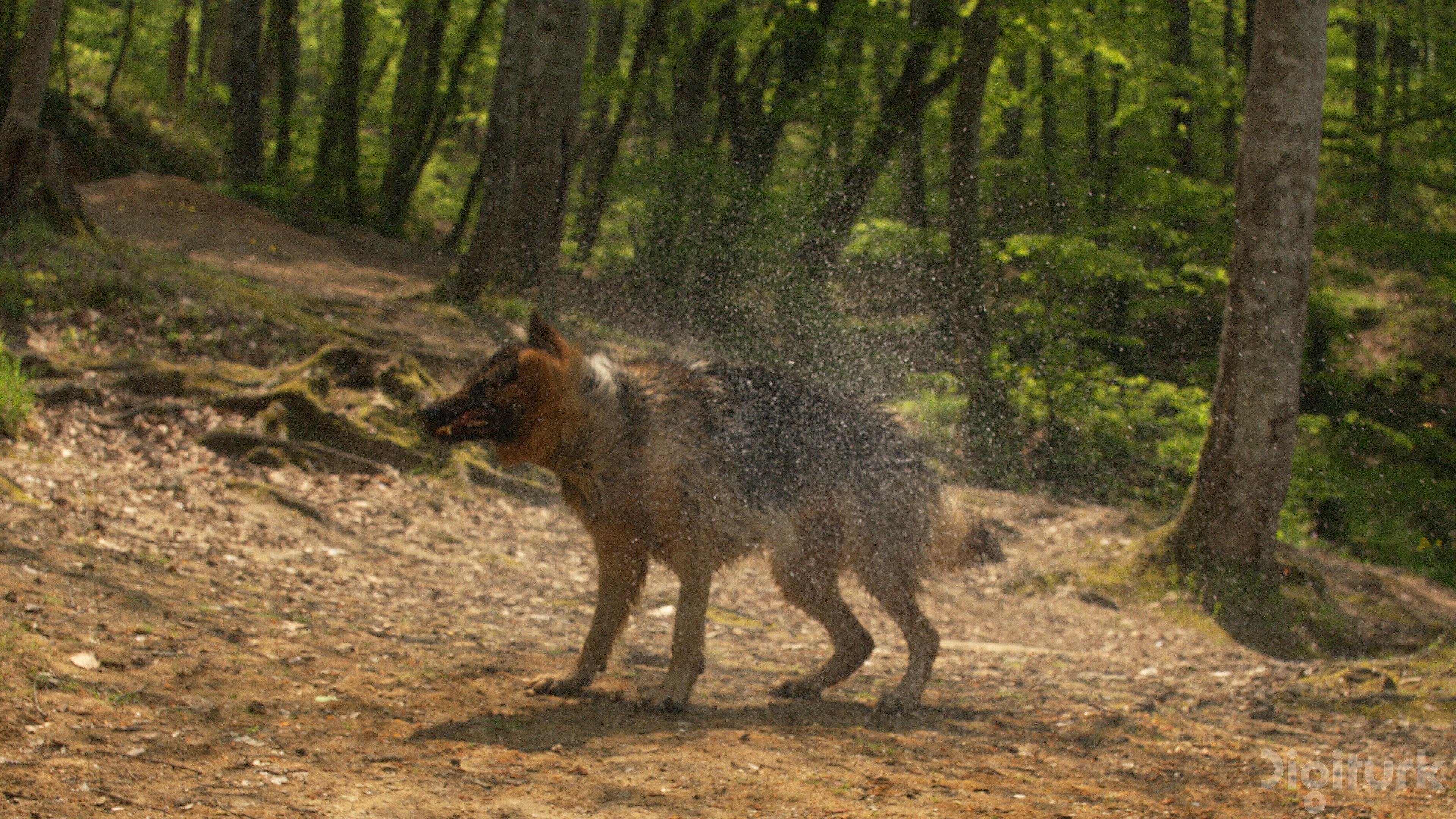}};
            \draw[red, thick] (-1.4,0.4) rectangle (-0.4,-0.3);
        \end{tikzpicture}
    \end{minipage}
    \begin{minipage}{0.45\textwidth}
        \begin{tikzpicture}
            \node[inner sep=0pt] (img) at (0,0) {\includegraphics[width=0.48\linewidth]{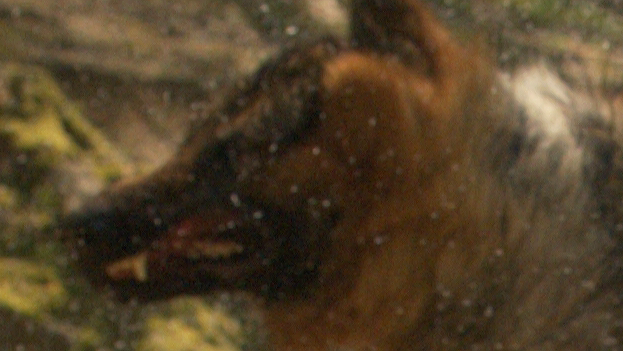}};
            \node[anchor=south east] at (img.south east) [fill=white, inner sep=1pt] {\tiny Ground Truth};
        \end{tikzpicture}%
        \begin{tikzpicture}
            \node[inner sep=0pt] (img) at (0,0) {\includegraphics[width=0.48\linewidth]{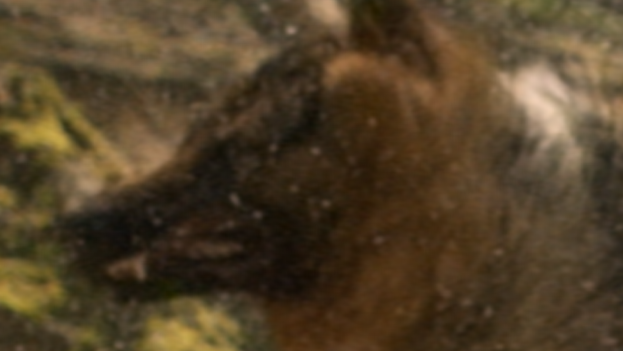}};
            \node[anchor=south east] at (img.south east) [fill=white, inner sep=1pt] {\tiny No Patching};
        \end{tikzpicture}\\[0.3em]
        \begin{tikzpicture}
            \node[inner sep=0pt] (img) at (0,0) {\includegraphics[width=0.48\linewidth]{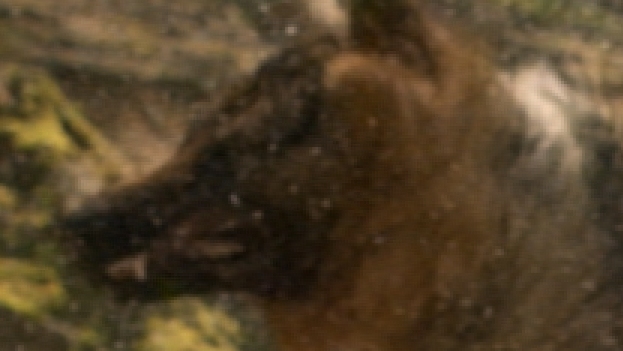}};
            \node[anchor=south east] at (img.south east) [fill=white, inner sep=1pt] {\tiny $3 \times 3$ Patch};
        \end{tikzpicture}%
        \begin{tikzpicture}
            \node[inner sep=0pt] (img) at (0,0) {\includegraphics[width=0.48\linewidth]{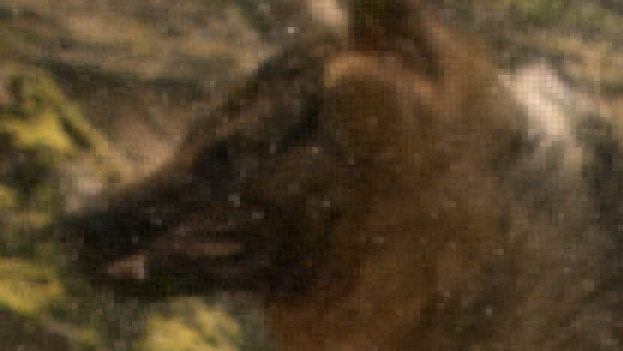}};
            \node[anchor=south east] at (img.south east) [fill=white, inner sep=1pt] {\tiny $6 \times 6$ Patch};
        \end{tikzpicture}
    \end{minipage}
    \caption{Visual comparison between the ground truth frame, patching output, and the default SIEDD-L model output on UVG-4k ShakeNDry. Pixels within a patch are always very similar, causing a pixelated effect.}
    \label{4kpatch}
\end{figure}

We also provide qualitative results from the UVG-4k experiments. We compare a ground truth frame from ShakeNDry with the result from SIEDD-L, SIEDD-L with a $3 \times 3$ patch, and SIEDD-L with a $6 \times 6$ patch in \ref{4kpatch}. While patching improved the decoding speed enormously, it comes at a cost of higher encoding time and a loss in reconstruction quality. The pixels within a patch are relatively uniform, causing an effect similar to nearest-neighbor upsampling.

\section{Additional Reconstruction Metrics}

\begin{figure}[t]
    \centering

    \begin{minipage}{0.45\textwidth}
        \begin{tikzpicture}
            \node[inner sep=0pt] (img) at (0,0) {\includegraphics[width=\linewidth]{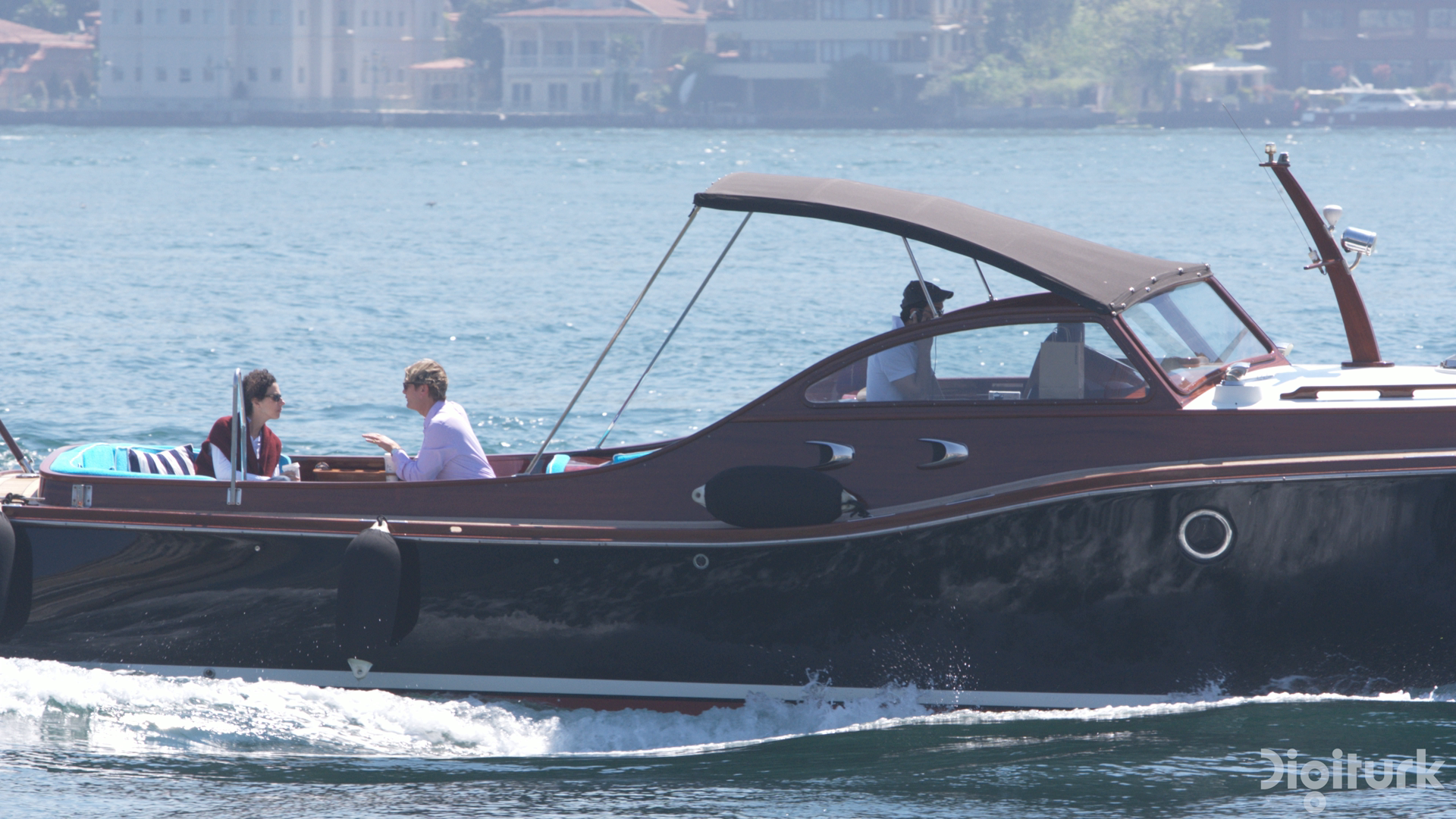}};
            \node[anchor=south east] at (img.south east) [fill=white, inner sep=1pt] {\tiny Ground Truth};
        \end{tikzpicture}
    \end{minipage}
    \begin{minipage}{0.45\textwidth}
        \begin{tikzpicture}
            \node[inner sep=0pt] (img) at (0,0) {\includegraphics[width=\linewidth]{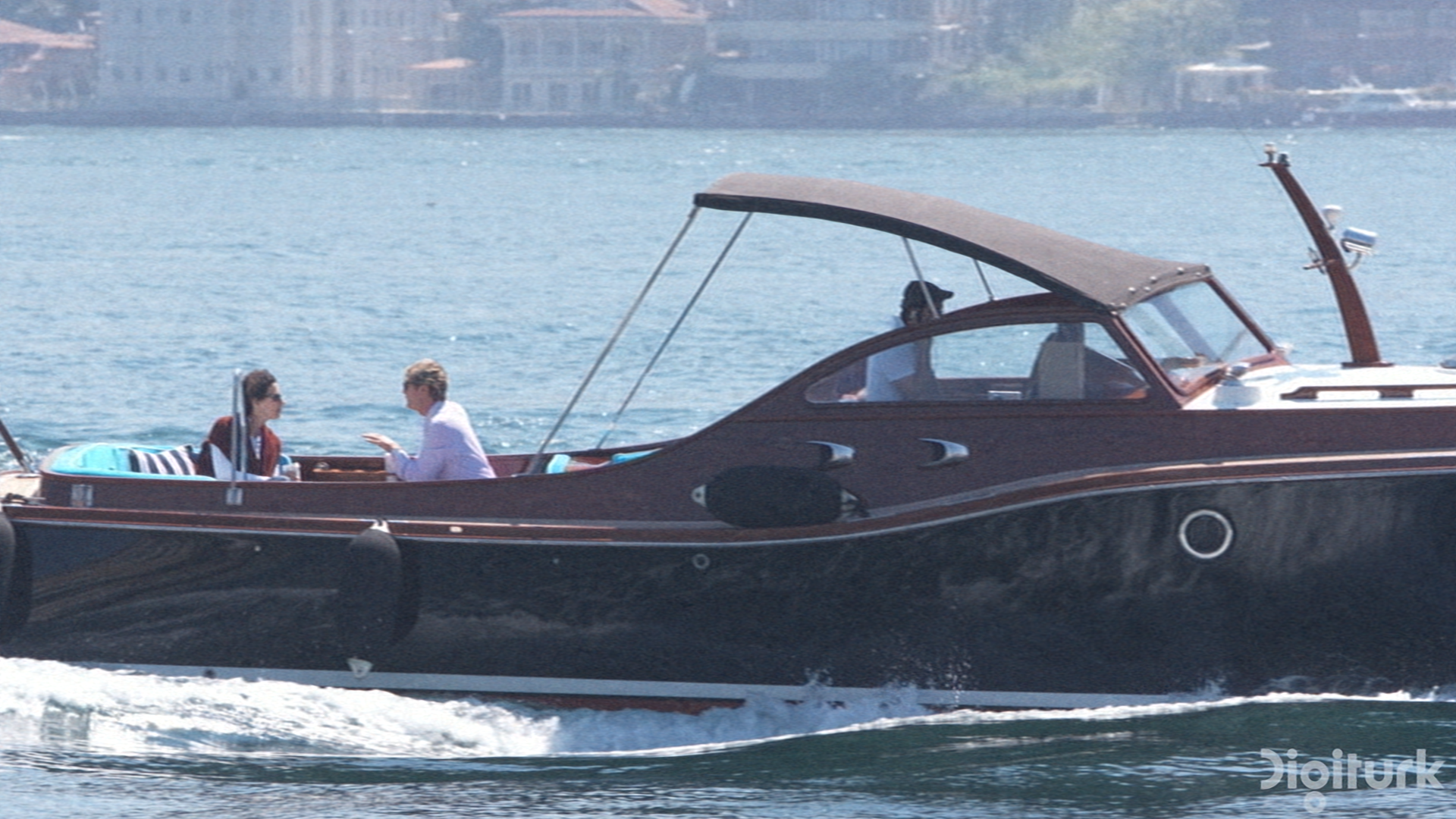}};
            \node[anchor=south east] at (img.south east) [fill=white, inner sep=1pt] {\tiny SIEDD-L Reconstruction};
        \end{tikzpicture}
    \end{minipage}
    \begin{minipage}{0.45\textwidth}
        \begin{tikzpicture}
            \node[inner sep=0pt] (img) at (0,0) {\includegraphics[width=\linewidth]{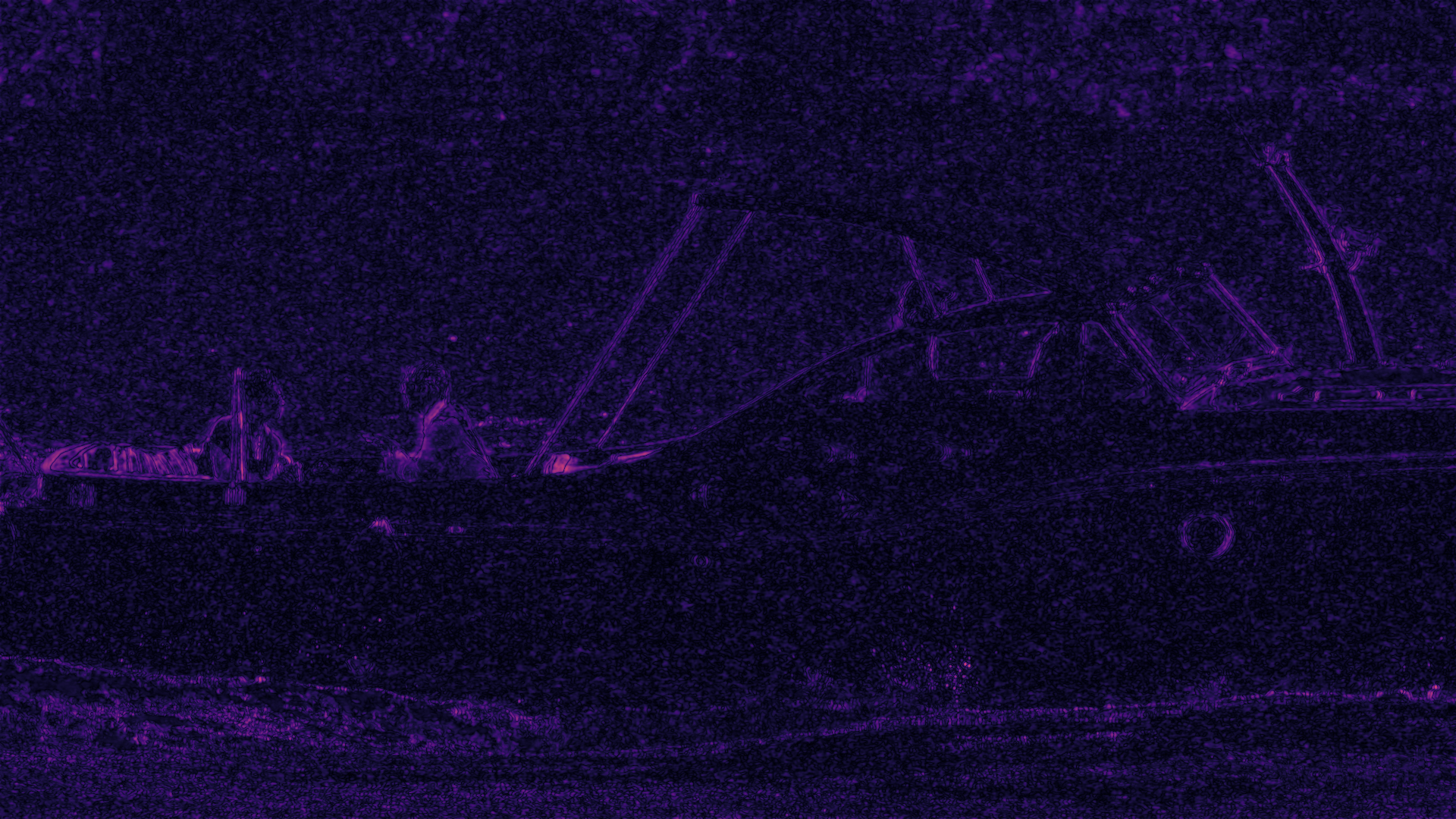}};
            \node[anchor=south east] at (img.south east) [fill=white, inner sep=1pt] {\tiny FLIP Map};
        \end{tikzpicture}
    \end{minipage}
    \caption{FLIP Visualization on UVG-HD YachtRide using SIEDD-L}
    \label{flip}
\end{figure}
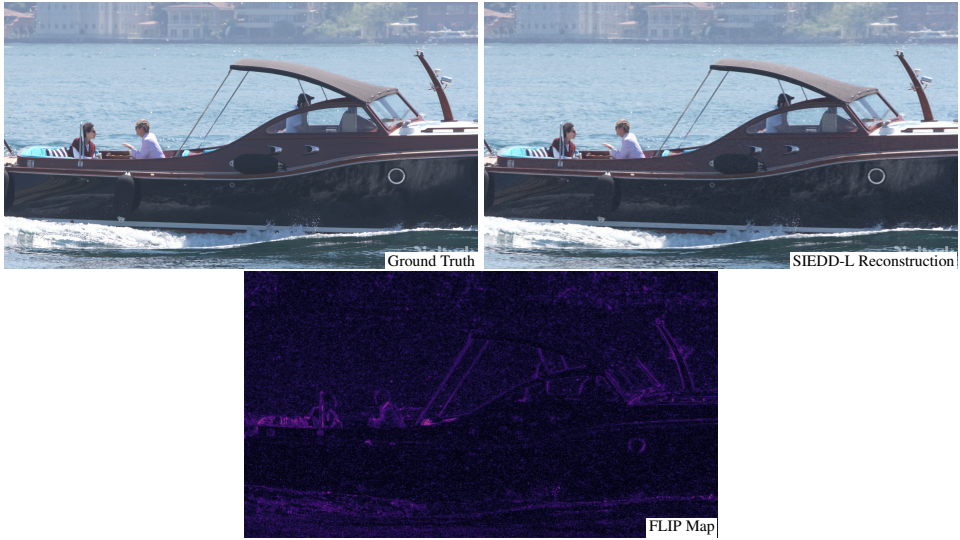

\begin{table}[!htp]\centering
\caption{VMAF and FLIP metrics on UVG-HD using SIEDD-S, SIEDD-M, and SIEDD-L}\label{tab:additionalmetrics}
\scriptsize
\begin{tabular}{lcccccccc}\toprule
Model & FLIP $\downarrow$ & VMAF $\uparrow$ \\\midrule
SIEDD-S & 0.100&66.69\\
SIEDD-M & 0.080 & 80.85\\
SIEDD-L & 0.068 & 87.75\\
\bottomrule
\end{tabular}
\end{table}
To provide additional metrics to ensure high quality video reconstruction, we use FLIP \citep{Andersson2020} and VMAF \citep{aistov2024vmafreimplementationpytorchexperimental}. We evaluate these metrics on SIEDD-S, SIEDD-M, and SIEDD-L on the UVG-HD dataset. We visualize the FLIP error map in \ref{flip} on a frame from UVG-HD YachtRide and list the FLIP and VMAF metrics in \ref{tab:additionalmetrics}.

\section{Effect of Group Size}

\begin{table}[t]\centering
\caption{Group Size Ablation}\label{tab:group-size}
\scriptsize
\begin{tabular}{lcccccccc}\toprule
Group Size &PSNR $\uparrow$ &SSIM $\uparrow$ &bpp $\downarrow$ &Time (s) $\downarrow$ \\\midrule
10 &36.54 &0.907 &0.587 &14915 \\
15 &35.67 &0.892 &0.402 &11171 \\
20 &34.84 &0.878 &0.297 &9296 \\
25 &34.52 &0.871 &0.255 &8720 \\
30 &34.14 &0.864 &0.218 &8225 \\
\bottomrule
\end{tabular}
\end{table}

To study the effect of the group size parameters ($N_g, N_s$) on SIEDD, we ran an ablation experiment on SIEDD-M in \ref{tab:group-size}. SIEDD-M uses $N_g=N_s=20$ by default, so in this experiment, we test the values of 10, 15, 25, and 30. We encode and evaluate the metrics of SIEDD-M with each group size on the UVG-HD dataset.

The group size hyperparameter is negatively correlated with the reconstruction quality (PSNR, SSIM) and compression (bpp) and is positively correlated with the encoding time. This is because with a larger group size, more frames share a single decoder, decreasing the number of parameters (and the bpp) for the video. A larger group size also means fewer decoder training loops, speeding up the encoding time. However, these come at a heavy cost to the reconstruction quality, and we chose 20 as the balanced default for SIEDD.

\section{Decoders with Low Rank Adaptation (LoRA)}

\begin{table}[!htp]\centering
\caption{LoRA Decoder Results}\label{tab:lora}
\scriptsize
\begin{tabular}{lcccccccc}\toprule
LoRA Type &LoRA Rank &PSNR $\uparrow$ &SSIM $\uparrow$ &bpp $\downarrow$ &fps $\uparrow$ &Time (s) $\downarrow$ \\\midrule
None (SIEDD-M) & - & 34.84&0.878&0.297& 6.27& 9296\\\midrule
Sin LoRA &1 &35.65 &0.891 &0.385 &6.30 &42802 \\
Sin LoRA &2 &35.78 &0.893 &0.432 &6.31 &42781 \\
Sin LoRA &4 &35.96 &0.896 &0.523 &6.36 &42614 \\
Sin LoRA &8 &36.22 &0.900 &0.703 &6.34 &43385 \\\midrule
LoRA &1 &35.58 &0.890 &0.385 &6.30 &41701 \\
LoRA &2 &35.71 &0.892 &0.431 &6.30 &42286 \\
LoRA &4 &35.87 &0.894 &0.523 &6.33 &42389 \\
LoRA &8 &36.10 &0.898 &0.700 &6.34 &42787 \\
\bottomrule
\end{tabular}
\end{table}

In SIEDD, the shared decoder has no independent parameters for each output frame with the exception of the last layer which produces the output coordinates. To introduce additional parameters that are unique to each video frame to the decoders, we use low rank adapters on each decoder layer.

In particular, we test traditional LoRA \citep{hu2021loralowrankadaptationlarge} and Sine LoRA \citep{ji2025efficientlearningsineactivatedlowrank}. Sin LoRA introduces a sine nonlinearity when multiplying the low rank matrices, allowing for more representative abilities. We find that the Sin LoRA provides marginal boosts in reconstruction quality (PSNR) compared to traditional LoRA.

Due to the materialization of a unique activation for each video frame in the forward pass, using LoRA adapters on the decoder layers increase encoding time by a significant value, as seen in \ref{tab:lora}.

\section{Quantization Results}

\begin{table}[t]\centering
\scriptsize
\begin{tabular}{lcccccccc}\toprule

Method & Bits & PSNR $\uparrow$ & SSIM $\uparrow$ & bpp $\downarrow$ & fps $\uparrow$ \\\midrule
None &32 &35.00 &0.880 &1.294 &6.38 \\\midrule
BNB &4 &30.19 &0.789 &0.184 &6.25 \\
BNB &8 &34.96 &0.879 &0.335 &3.09 \\\midrule
HQQ &4 &32.54 &0.838 &0.222 &6.26 \\
HQQ &6 &34.97 &0.879 &0.310 &6.26 \\
HQQ &8 &35.16 &0.882 &0.378 &6.26 \\\midrule
Post & 4 &15.55 &0.418 &0.178 &6.39 \\
Post & 5 &27.34 &0.758 &0.220 &6.38 \\
Post & 6 &33.13 &0.852 &0.263 &6.39 \\
Post & 7 &34.51 &0.873 &0.306 &6.38 \\
Post & 8 &34.87 &0.878 &0.348 &6.37 \\
\bottomrule
\end{tabular}
\caption{Quantization Sweep on SIEDD-M with Different Methods}\label{tab:quantsweep-methods}
\end{table}

\begin{table}[t]\centering

\begin{minipage}[t]{0.45\textwidth}
\centering
\scriptsize

\begin{tabular}{lcccc}\toprule
Bit &bpp $\downarrow$ &PSNR $\uparrow$ &SSIM $\uparrow$ \\\midrule
4 &0.106 &31.19 &0.812 \\
5 &0.127 &32.78 &0.841 \\
6 &0.145 &33.24 &0.849 \\
8 &0.179 &33.39 &0.851 \\
\bottomrule
\end{tabular}
\subcaption{SIEDD-S}\label{tab:quantsweep-s}
\end{minipage}
\begin{minipage}[t]{0.45\textwidth}
\centering
\scriptsize
\begin{tabular}{lcccc}\toprule
Bit &bpp $\downarrow$ &PSNR $\uparrow$ &SSIM $\uparrow$ \\\midrule
4 &0.222 &32.82 &0.843 \\
5 &0.268 &34.50 &0.872 \\
6 &0.310 &35.00 &0.880 \\
8 &0.377 &35.16 &0.882 \\
\bottomrule
\end{tabular}
\subcaption{SIEDD-M}\label{tab:quantsweep-m}
\end{minipage}
\begin{minipage}[t]{0.45\textwidth}
\centering
\scriptsize
\begin{tabular}{lcccc}\toprule
Bit &bpp $\downarrow$ &PSNR $\uparrow$ &SSIM $\uparrow$ \\\midrule
4 &0.380 &34.00 &0.865 \\
5 &0.461 &35.77 &0.895 \\
6 &0.534 &36.31 &0.903 \\
8 &0.644 &36.49 &0.905 \\
\bottomrule
\end{tabular}
\subcaption{SIEDD-L}\label{tab:quantsweep-l}
\end{minipage}
\caption{Quantization Sweep for HQQ on SIEDD-S, SIEDD-M, and SIEDD-L}
\label{tab:quantsweep}
\end{table}

Due to the importance of quantization in our compression pipeline, we experimented with different quantization methods. Traditional post-training quantization (PTQ) involves scaling the tensor from 0 to $2^b$ where b is the number of bits to quantize to, casting this tensor to an integer, and storing it. While this method works for larger $b$ values, performance degrades rapidly at lower precision integers as shown in \ref{tab:quantsweep-methods}. We also show the performance of BNB \citep{dettmers2022llmint8} which has 4 bit and 8 bit implementations. However, the best performing method at low precision is hqq \citep{badri2023hqq} which shows minimal degregation in PSNR and SSIM compared to other methods at low bpp.

To further show the limits of hqq in the SIEDD architecture, we tested the reconstruction quality against bpp for each of the SIEDD-S, SIEDD-M, and SIEDD-L models in \ref{tab:quantsweep}. As 6 bit gave a negligible (<0.2 PSNR) performance drop compared to 8 bit, this became the baseline for the SIEDD models.

\section{Decoding}

Decoding a coordinate based model such as SIEDD is an extremely computationally intensive task. A naive implementation will result in out-of-memory errors and a low FPS. To make 1080p and 4k decoding possible, the forward pass must be batched in terms of coordinates and the separate decoders. In our implementation, we chunk the coordinates into 8 separate forward passes, and within each forward pass, the decoders are run one at a time. For 4k decoding, we instead chunk the coordinates into 32 forward passes of the model, ensuring that 4k videos can be encoded using an A5000 GPU. In addition, the SIEDD model parameters are converted to bfloat16 to accelerate decoding speed. To calculate the decoding speed (fps), we record the time required to run the forward pass on all coordinates on the GPU and send the frames to the CPU.

For our LoRA experiments, the FPS would normally be extremely low due to the materialization of activations for each individual frame, as opposed to one activation for the frame group. As a result, LoRA decoding performance would have been even worse. However, we split the shared decoder and LoRA adapters into separate decoder layers, effectively splitting the decoder into $N_g$ parallel MLPs. This allows the FPS to be on par with SIEDD without LoRA adapters.

\section{Video Denoising}
\begin{table}[!htp]\centering
\caption{Video Denoising PSNR Results}\label{tab:denoising}
\scriptsize
\begin{tabular}{lcccccccc}\toprule
&All White &All Black &Salt \& Pepper & Random \\\midrule
Baseline &28.08 &29.69 &28.07 &30.95 \\
Gaussian Blur &35.25 &36.03 &35.56 &36.71 \\
Median Blur &36.47 &36.46 &36.47 &36.47 \\
SIEDD-L &35.66 &35.66 &35.66 &35.66 \\
\bottomrule
\end{tabular}
\end{table}
We also showcase SIEDD's ability to carry out image denoising in \ref{tab:denoising}. To achieve this, we add different types of image noise to UVG-HD, encode the video using traditional denoising methods and SIEDD, and compare the output quality to the original frames. To add noise, we tested setting certain pixels to all white, all black, adding salt and pepper noise, and random noise. We added noise to $10^4$ pixels of each frame using these different methods. The traditional denoising methods include gaussian and median blurs. We also record the baseline which is the PSNR of the noisy image and the ground truth. The results show a comparable performance to traditional denoising techniques such as gaussian and median blurring with the added benefits of SIEDD.


\bibliographystyle{unsrtnat}
\bibliography{citations}

\end{document}